\documentclass[pdflatex,sn-nature]{sn-jnl}
\usepackage{graphicx}%
\usepackage{multirow}%
\usepackage{amsmath,amssymb,amsfonts}%
\usepackage{amsthm}%
\usepackage{mathrsfs}%
\usepackage[title]{appendix}%
\usepackage{xcolor}%
\usepackage{textcomp}%
\usepackage{manyfoot}%
\usepackage{booktabs}%
\usepackage{algorithm}%
\usepackage{nicefrac}  
\usepackage{algorithmicx}%
\usepackage{algpseudocode}%
\usepackage{listings}%
\usepackage{ifthen}
\usepackage{booktabs, cellspace, hhline}
\usepackage{xr}

\usepackage{adjustbox}
\newcommand{\ourmodel}{OrbitAll}
\newcommand{\foundationmodel}{OrbitAll-OMol25-4M}
\newcommand{\tlxsolv}{T1x-Solv}

\newboolean{arxiv}
\setboolean{arxiv}{true} % set it true if submitting to arxiv}
\newboolean{markdiff}
\setboolean{markdiff}{false}

\ifthenelse{\boolean{arxiv}}
{
\newcommand{\SI}{}
\newcommand{\appendixsection}{Appendix}
}
{
\externaldocument{si}
\newcommand{\SI}{SI }
\newcommand{\appendixsection}{Section}
}

\ifthenelse{\boolean{markdiff}}
{
\newcommand{\markdiff}[1]{\textcolor{red}{#1}}
}
{
\newcommand{\markdiff}[1]{#1}
}

\theoremstyle{thmstyleone}

\theoremstyle{thmstyletwo}%

\theoremstyle{thmstylethree}%

\begin{document}

\title[\ourmodel{}: A Unified Quantum Mechanical Representation Deep Learning Framework for All Molecular Systems]{\ourmodel{}: A Unified Quantum Mechanical Representation Deep Learning Framework for\\All Molecular Systems}

\author[1]{\fnm{Beom Seok} \sur{Kang}}
\equalcont{These authors contributed equally to this work.}

\author[1]{\fnm{Vignesh C.} \sur{Bhethanabotla}}
\equalcont{These authors contributed equally to this work.}

\author[2]{\fnm{Amin} \sur{Tavakoli}}

\author[2]{\fnm{Maurice D.} \sur{Hanisch}}

\author[2]{\fnm{Arimitsu} \sur{Horikawa-Strakovsky}}

\author[1]{\fnm{Miguel} \sur{Nouman}}

\author[2]{\fnm{Danish} \sur{Khan}}

\author*[1]{\fnm{William A.} \sur{Goddard III}}\email{wag@caltech.edu}

\author*[2]{\fnm{Anima} \sur{Anandkumar}}\email{anima@caltech.edu}

\affil[1]{\orgdiv{Division of Chemistry and Chemical Engineering}, \orgname{Caltech}, \orgaddress{\street{1200 E California Blvd}, \city{Pasadena}, \postcode{91125}, \state{CA}, \country{USA}}}

\affil[2]{\orgdiv{Division of Engineering and Applied Science}, \orgname{Caltech}, \orgaddress{\street{1200 E California Blvd}, \city{Pasadena}, \postcode{91125}, \state{CA}, \country{USA}}}

\abstract{%Deep learning methods are often designed for restricted chemical settings, whereas real-world systems can vary in charge, spin, and environment. 
\markdiff{We introduce \ourmodel{}, a geometry- and physics-informed deep learning framework that encodes any molecular system with arbitrary charges, spins, and environmental effects using electronic structure information. It utilizes spin-polarized orbital features from the underlying quantum mechanical method and combines them with SE(3)-equivariant graph neural networks. \ourmodel{} demonstrates superior performance and generalization in predicting charged, open-shell, and solvated molecules, and robustly extrapolates to molecules significantly larger than the training data. \ourmodel{} achieves chemical accuracy using 10 times fewer training data than competing AI models, with approximately 10\textsuperscript{3} -- 10\textsuperscript{4} speedup compared to density functional theory. Trained on a chemically diverse dataset, \ourmodel{} performs robustly on challenging molecular systems, and outperforms the foundational machine-learned interatomic potential, UMA, for highly charged species, despite using 35 times less molecular data and a 50-times-smaller model. After learning solvent effects, it accurately predicts solvent-dependent reaction pathways at about 100 times lower cost than explicit-solvation simulations using UMA.}}


\maketitle

\section{Introduction}\label{sec:intro}

Quantum mechanical (QM) methods, such as density functional theory (DFT), can accurately simulate atomic and molecular systems \cite{cohen2012challenges, pereira2017machine, mood2020methyl}, but require an enormous computational cost \cite{frison_compare_dft&semi_2008, engel2011dft, friesner2005abinitio}. 
Machine learning (ML)-based methods enable high-throughput atomic simulations at a fraction of the computational cost, while maintaining relative quantum chemical accuracy \cite{fedik_ml4molprops_2022, fooshee2018deep,musaelian2023allegro}. These ML models, after training on relevant datasets, can be orders of magnitude faster than traditional quantum mechanical simulations while still delivering reasonably accurate predictions \cite{musaelian2023allegro, qiao_orbnet_2020, christensen2021orbnetdenali, batzner2022nequip, tavakoli2025chemically, tavakoli2022quantum}. However, the use of these ML models is often limited to the chemistry of their training data.

\markdiff{Foundation models for machine-learned interatomic potentials (MLIPs), such as UMA \cite{wood2025uma}, have demonstrated strong accuracy across a broad range of molecular systems, including those with varying charge and spin states, but with crucial limitations. First, the predictive performance of these models is mainly bounded by the chemical domains represented in their training data. Their strong generalization therefore reflects interpolation across sufficiently diverse training distributions, rather than a principled ability to extrapolate to regimes that are sparsely sampled or absent from the data. As a result, such models may fail in the extreme or unusual cases where physical guidance is most needed. Second, current models often do not natively incorporate environmental effects, such as implicit solvation or external electric fields, which can substantially alter molecular electronic structure and reactivity.}

Realistic application to all chemical systems with a generalization capability on par with QM calculations requires taking additional degrees of freedom into account: spin, charge, and environmental effects \cite{faglioni2016battery2, das2024battery3}. 
Integrating spin effects enables applicability to open-shell systems, where unpaired electrons significantly impact the electronic structure and reaction pathways \cite{borden_diradical_2017, tavakoli2023rmechdb, tavakoli2024ai, kang2024geometry, kang2024orbnetspin}. For example, transition metal complexes and radicals require open-shell calculations to capture spin-related effects accurately \cite{maurer2021organometallic, gutowski2013radicals}.
Integrating charge effects allows for the accurate modeling of charged species, which are essential in many chemical systems \cite{he2019dftbattery1}. In biochemistry, charged states play a crucial role in enzyme catalysis and protein-ligand interactions, where protonation states and ionic interactions influence binding affinities and reaction mechanisms \cite{dudev2014competition, roberts2015specific}.
Further, integrating environmental effects, such as solvation and external electric fields, enhances the modeling of molecular interactions under realistic conditions. These factors can alter electronic properties and reaction energetics, influencing processes such as molecular solvation and field-assisted catalysis \cite{liang2017efield, zhao2011solvation, wei2000solvation}.

Incorporating QM information into molecular representations enhances molecular modeling, resulting in highly accurate predictive models.  \cite{qiao_orbnet_2020, qiao_orbnet_equi_2022, karandashev2022qml, cheng2019mobml1, lu2022mobml2, cheng2019mobml3, cheng2022mobml4, cheng_MOBML_2022, venturella2025mlgf}. Compared to purely data-driven or even geometry-based models, QM-informed models offer improved generalization and data-efficiency, mitigating the high computational cost of data generation for training predictive models \cite{qiao_orbnet_equi_2022, rama_delta_learning_2015, karandashev2022qml, fooshee2018deep, irwin2022chemformer}. For example, OrbNet achieved chemical accuracy on relative conformer energies for molecules much larger than those of the training set, demonstrating strong size extrapolation \cite{qiao_orbnet_2020}. Furthermore, leveraging QM-informed features—a strategy we call \textit{orbital learning}—enables models to more effectively capture underlying physics by implicitly encoding spin, charge, and environmental effects \cite{qiao_orbnet_equi_2022, cheng2019mobml1, fabrizio2022spa, briling2024spahm, venturella2025mlgf}.

\begin{figure}[!ht]
\begin{center}
\centerline{\includegraphics[width=1.0\columnwidth]{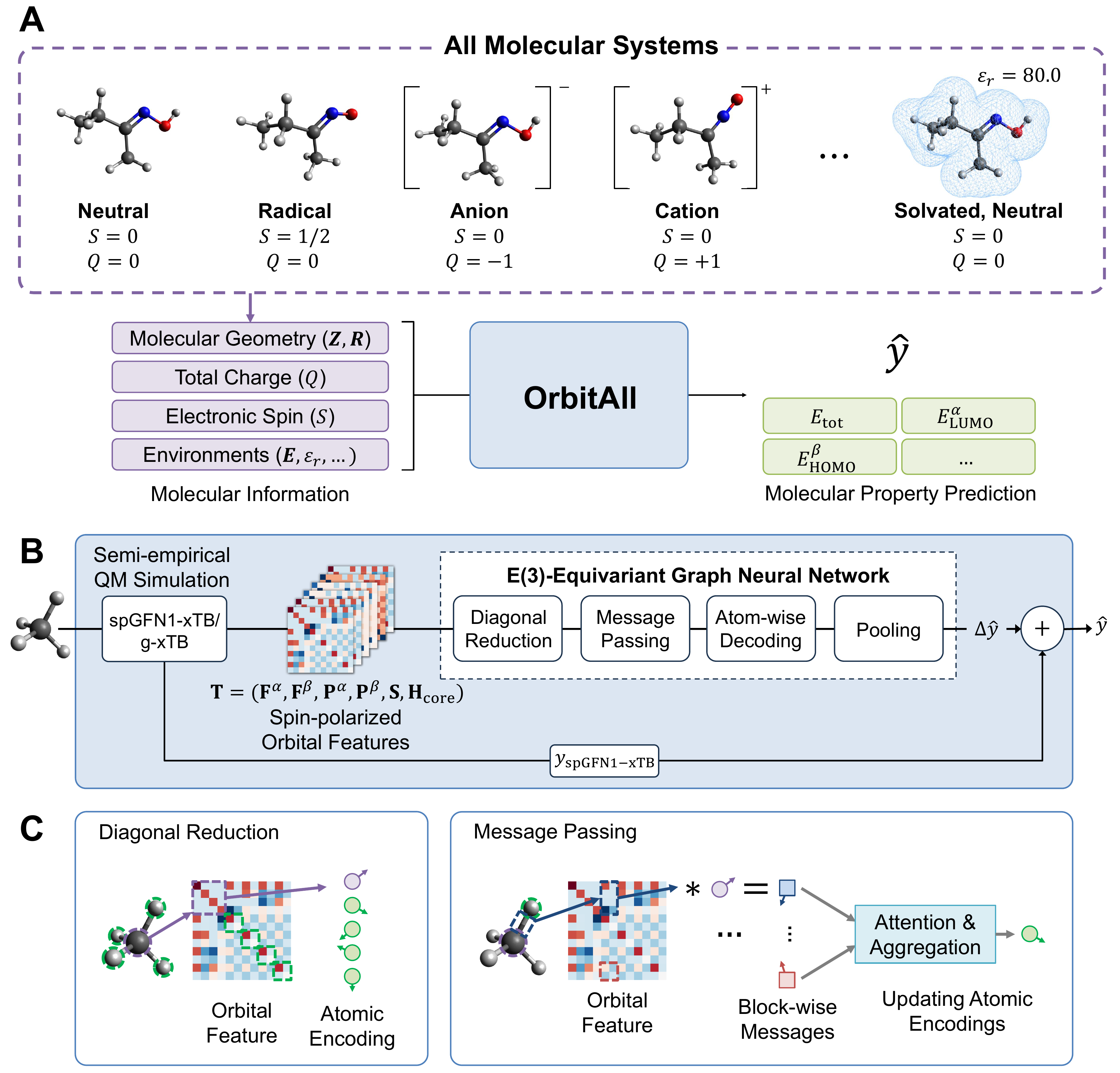}}

\caption{\textbf{Schematic of the \ourmodel{} framework.} \textit{(A)} Illustration of inputs and outputs. \ourmodel{} can process all possible molecular systems for quantum mechanical calculations with different total spin, total charge, and environments. With this molecular system information, \ourmodel{} predicts molecular properties. \textit{(B)} Overall architecture of \ourmodel{}. First, the molecular system's information is processed by a semi-empirical QM method, spGFN1-xTB \cite{neugebauer_spgfnxtb_2023}, to create the spin-polarized orbital features set, $\textbf{T}=(\textbf{F}^\alpha,\textbf{F}^\beta,\textbf{P}^\alpha,\textbf{P}^\beta,\textbf{S},\textbf{H}_\text{core})$. This set of orbital features can represent all molecular systems for quantum mechanical simulations. Second, the orbital features, which are SE(3)-equivariant, are processed by an E(3)-equivariant GNN. The target we predict with the GNN is the delta-label \cite{rama_delta_learning_2015}, $\Delta y$, the difference between the low-level approximation obtained from spGFN1-xTB, $y_{\text{spGFN1-xTB}}$, and the high-level label, $y$. \textit{(C)} Illustrations of the diagonal reduction and the message passing operations using the QMMs. The diagonal reduction encodes the block-diagonals of QMMs into atom-wise encodings (representations). The messages are created by convolutions of the atom-wise representations with the off-block diagonals, then are aggregated with attention to update the atom-wise representations. Further details can be found in Section \ref{sec:methods}.}

\label{fig:overview}
\end{center}
\end{figure}
% \newpage

{\bf Our approach:} We introduce ``\ourmodel{}'', a geometry- and physics-informed deep learning framework for representing all molecular systems with arbitrary charge, spin, and environmental effects (Figure \ref{fig:overview}). \ourmodel{} begins by representing electronic structure using a low-cost semi-empirical QM method to produce SE(3)-equivariant orbital features that encapsulate converged mean-field electronic interactions. These quantum mechanical features are then mapped onto atomic graphs to retain both geometric and electronic information. Then, an E(3)-equivariant graph neural network (GNN) with orbital-based message-passing processes these features to predict target properties. Therefore, both the representations and the neural network are grounded in orbital features, closely aligning with the underlying physics and consequently enhancing generalization and extrapolation capability.

\ourmodel{} is the first orbital-based deep learning framework that uses a single unified representation to jointly model all molecular systems across varying charge, spin, and environmental conditions. In contrast, most state-of-the-art geometric GNNs support only a subset of these features, often requiring separate extensions or input modifications to address them individually.

% \markdiff{We train \ourmodel{} on the chemically diverse \aacomment{no you don't. you train on a subset of it. can u state precisely how big and which subset} OMol25 dataset~\cite{levine2025openmolecules2025omol25}, yielding a general-purpose model, \foundationmodel{}, that achieves performance competitive with state-of-the-art machine-learned interatomic potentials (MLIPs). \aacomment{Although trained on a smaller subset and smaller in size compared to popular models like UMA, ...}Notably, \foundationmodel{} remains robust for complex and underrepresented species, including highly charged ions, and outperforms substantially larger-scale models such as UMA on challenging subsets~\cite{wood2025uma}.}

\markdiff{We conduct a large-scale training of \ourmodel{} for a general-purpose model, trained on the OMol25-4M dataset~\cite{levine2025openmolecules2025omol25}. This dataset is a 4-million-molecule subset of the larger OMol25 dataset, which consists of 140 million molecules and was used to train the UMA foundation model, the state-of-the-art MLIP~\cite{wood2025uma}. Although \ourmodel{} is trained on a substantially smaller dataset (35 times smaller for molecular data and 130 times smaller for the total) and is smaller in size (50 times smaller) compared to UMA, it remains robust for complex and underrepresented species, including highly charged ions, and outperforms UMA on challenging subsets despite.}

%, \tlxsolv{},

\markdiff{We fine-tune \ourmodel{} on an implicit-solvation dataset to construct a solvent-effect-informed interatomic potential. The unified representation, which enables the incorporation of field effects, allows \ourmodel{} to process molecules under implicit solvation in different solvents, an approach not directly applicable to conventional MLIPs such as UMA, and those models require expensive explicit solvation to account for solvent effects. Using the nudged elastic band method, \ourmodel{} accurately captures solvent-dependent transition-state structures and reaction energetics at about 100 times lower cost than explicit-solvation umbrella sampling with UMA.}

We further do ablations and evaluations of the performance of energy prediction tasks using carefully selected datasets, including QM9star (for varying spins and charges) \cite{tang2024qm9star} and Hessian QM9 (for varying implicit solvents) \cite{williams2025hessianqm9}.
We first compare model accuracy across various molecular property predictions, showing that \ourmodel{} consistently outperforms all augmented models in total energy and frontier orbital energy estimation. Next, the cost-accuracy tradeoff and data-efficiency of the models are compared. The result reveals that \ourmodel{} achieves a mean absolute error below chemical accuracy (1 kcal/mol) using only 10\% of the data required to train the next-best model, while being $\sim10^3-10^4$ faster than DFT/B3LYP during inference.

We also examine the generalization capability of the models to molecular systems of much larger sizes using a dataset of polypeptides. We observe that \ourmodel{} robustly outperforms competing models for these much larger molecules. Additionally, we demonstrate \ourmodel{}'s ability to predict the molecular energies for various implicit solvents using Hessian QM9, which again exhibits remarkable generalization across different solvent environments.

\markdiff{Summarizing, \ourmodel{} overcomes many crucial limitations of current state-of-the-art foundational MLIPs, which remain constrained by the chemical domains represented in their training data. Furthermore, \ourmodel{} can natively incorporate environmental effects, such as implicit solvation or external electric fields, that are missed by current MLIPs, which have substantial influence on molecular electronic structure and reactivity.}

\section{Results}
\label{sec:results}

\subsection{The OrbitAll Framework}

Incorporating molecular geometry into predictive models significantly improves the accuracy of molecular property prediction \cite{schutt2018schnet, Gasteiger_dimenet_2020}. Moreover, designing models that respect molecular symmetries—through group equivariance—further enhances both accuracy and data efficiency \cite{thomas2018tensor, batzner2022nequip}. Notable improvement is also achieved by orbital learning, where the use of physics-informed features (orbital features) enhances the accuracy and fidelity of predictions \cite{qiao_orbnet_2020, qiao_orbnet_equi_2022, cheng_MOBML_2022, karandashev2022qml}. The \ourmodel{} framework extends these methods to spin-polarizable systems, which enables processing all molecular systems.
As shown in Figure \ref{fig:overview}, the \ourmodel{} framework accepts as inputs atomic numbers ($\textbf{\textit{Z}}$), atomic coordinates ($\textbf{\textit{R}}$), total charge ($Q$), total spin ($S$), and parameters regarding environmental effects, such as implicit solvation. The spin-polarized orbital features, $\textbf{T}=(\textbf{F}^\alpha,\textbf{F}^\beta,\textbf{P}^\alpha,\textbf{P}^\beta,\textbf{S},\textbf{H}_\text{core})$, are generated using spGFN1-xTB \cite{neugebauer_spgfnxtb_2023} or g-xTB \cite{froitzheim2025gxtb}. Here, $\textbf{F}^\alpha$ and $\textbf{F}^\beta$ denote the Fock matrices for up-spin ($\alpha$) and down-spin ($\beta$), respectively; $\textbf{P}^\alpha$ and $\textbf{P}^\beta$ are the corresponding density matrices; $\textbf{S}$ is the overlap matrix; and $\textbf{H}_\text{core}$ is the core Hamiltonian matrix. We refer to these as quantum mechanical matrices (QMMs), which collectively represent the converged mean-field electronic structure of the molecular system.

The orbital features that represent the electronic structure of the input system are perturbed correspondingly to the various conditions, since the low-level QM method (e.g., spGFN1-xTB) can physically capture the major effects.
Thus, this representation can distinguish molecules with different spin states (singlets, doublets, triplets, etc.), different charges (neutral, anion, cation), and different environmental effects (uniform external electric fields, various implicit solvents, etc.) in a common representation space. 

In this study, the set of orbital features used are $\textbf{T}=(\textbf{F}^\alpha,\textbf{F}^\beta,\textbf{P}^\alpha,\textbf{P}^\beta,\textbf{S},\textbf{H}_\text{core})$. Each element of a QMM $\textbf{O}$ generated from an operator $\hat{\mathcal{O}}$ is obtained by,
\begin{equation}
\label{eqn:element_qcm}
    (\textbf{O})^{n,l,m;n',l',m'}_{AB}=\langle\Phi^{n,l,m}_A|\hat{\mathcal{O}}|\Phi^{n',l',m'}_B\rangle,
\end{equation}

\noindent
where $n$, $l$, and $m$ are the principal, angular,  and magnetic quantum numbers, respectively \cite{szabo_quantum_1989}.  

\ourmodel{} employs an E(3)-equivariant neural network framework, UNiTE \cite{qiao_orbnet_equi_2022}, which serves as the backbone of the geometric GNN module. \ourmodel{} adheres to SE(3)-equivariance overall based on the predictable change of QMM elements, which are SE(3)-equivariant. For a roto-translational transformation $\mathcal{R}$,

\begin{equation}
    \label{eqn:qcm_block_rotation}
    (\mathcal{R}\cdot\textbf{O})^{l;l'}_{AB}=\textbf{D}^{l}(\mathcal{R})(\textbf{O})^{l;l'}_{AB}\left(\textbf{D}^{l'}(\mathcal{R})\right)^{\dagger},
\end{equation}

\noindent
where $\textbf{O}^{l;l'}_{AB}$ is a block in the QMM $\textbf{O}$ that represents the interaction between the atomic orbital of atom $A$ with an angular momentum $l$ and the atomic orbital of atom $B$ with an angular momentum $l'$. The dagger symbol denotes the Hermitian conjugate, and $\textbf{D}^l(\mathcal{R})$ is the Wigner-D matrix of degree $l$, an irreducible representation of SO(3).

Within the \ourmodel{} framework, the self-consistent field (SCF) method used to create orbital features yields the quantum mechanical properties at that level of theory without any additional cost. This low-level approximation enables delta-learning (or $\Delta$-learning, $\Delta$-ML) strategy \cite{rama_delta_learning_2015}, where the task changes into predicting the difference of the original label (higher-level theory or experimental) from the lower-level approximation. With a delta-learning objective, the ML model predicts the delta-label $\Delta y$, defined by:

\begin{equation}
    \Delta y=y_\text{target}-y_\text{low-level},
\end{equation}

\noindent
where $y_\text{target}$ is the target label for prediction and $y_\text{low-level}$ is the low-level approximation (e.g., spGFN1-xTB) of the quantity. The delta-learning strategy has been empirically shown to improve the accuracy and data-efficiency of several different labels, since the low-level approximation can capture essential electronic interactions and the resulting delta energy surface is easier to learn \cite{qiao_orbnet_equi_2022, rama_delta_learning_2015, ruth2022deltalearning1, chen2023deltalearning2, zhu2019deltalearning4}.
Here, for comparison, we refer to direct predictions of target labels as direct-learning to highlight the differences with delta-learning.

\subsection{\markdiff{Diverse Chemistry and Real World Applications}}
\label{sec:foundation_model}

\paragraph{\markdiff{Large-Scale \ourmodel{} for Diverse Chemistry}}

%However, their predictive scope remains largely constrained to the domains represented in their training data. In particular, extending these models to capture environmental effects, such as implicit solvation or external electric fields, is nontrivial, as their architectures are not explicitly designed to incorporate such effects.

\markdiff{We train a large-scale \ourmodel{} as a QM-informed molecular foundational model on the OMol25 dataset \cite{levine2025openmolecules2025omol25}, which contains approximately 4 million single-point hybrid DFT calculations at the $\omega$B97M-V/def2-TZVPD level of theory. The dataset spans chemically diverse systems, various elements, highly non-equilibrium structures, and complex chemistries such as transition metal complexes. The resulting model, \foundationmodel{}, is trained using orbital features generated by g-xTB v2.0.0 as the internal QM method \cite{froitzheim2025gxtb}, with g-xTB energies and forces used as the delta-learning baseline.}

% To address this gap, we train the \foundationmodel{} model on the OMol25 dataset \cite{levine2025openmolecules2025omol25}, which contains approximately 4 million single-point hybrid DFT calculations ($\omega$B97M-V/def2-TZVPD). The dataset includes chemically diverse systems, various elements, highly non-equilibrium structures, and complex chemistries such as transition metal complexes. The \foundationmodel{} is trained using orbital features generated with g-xTB v2.0.0 as the internal QM method \cite{froitzheim2025gxtb}, with g-xTB energies and forces used as the delta-learning baseline. The distinct basis-set strategy used in g-xTB requires a new diagonal reduction algorithm. Additionally, we note that the lanthanides are removed from the datasets due to spin state ambiguity from using g-xTB as the QM method. Further details are provided in Section \ref{sec:methods}.

\markdiff{\foundationmodel{} achieves an MAE of 62.72 meV ($\sim$1.45 kcal/mol) and a per-atom MAE of 1.10 meV/atom on the evaluation set, comparable to those reported for state-of-the-art benchmark models \cite{levine2025openmolecules2025omol25}. The model also shows strong generalization across different chemical domains, including biomolecules, electrolytes, and metal complexes (\SI{}\appendixsection{} \ref{sec:app:omol25_result_additional}). }

%However, it exhibits a relatively higher force MAE of 20.64 meV/\AA, which may reflect limitations of the current backbone architecture in capturing force-sensitive local geometric variations.

%Additionally, the \foundationmodel{} model shows promising behavior and provides important insights. 

\markdiff{Most importantly, \foundationmodel{} performs robustly in extreme and underrepresented charged regimes. It achieves substantially lower MAEs than UMA-s-1p2 for ions with charges of +5 and above or -4 and below, except at -10 (\SI{}Figure \ref{fig:app:omol_eval_by_charge}), despite a substantially smaller training set: 35-fold smaller for finite molecules and 130-fold smaller overall. Such robustness is particularly valuable for reactive and chemically diverse systems, where unusual charge states can be important.}

%Most importantly, \foundationmodel{} remains superior in extreme and underrepresented regimes of charged species, . \foundationmodel{} records substantially lower MAEs than UMA-s-1p2 for ions of charges of +5 and above, or -4 and below, except -10 (\SI{}Figure \ref{fig:app:omol_eval_by_charge}). This generalization beyond training data is achieved despite the substantially smaller training set (35 times smaller for finite molecules, 130 times smaller for entire). This robustness could be particularly valuable for modeling reactive and chemically diverse systems, where unusual charge states may appear and play an important role.

%While minimum-energy pathway (MEP) construction is a key MLIP application, addressing solvent effects with MLIPs often requires explicit solvation, which greatly increases the system size and makes NEB pathways configuration-dependent. Implicit solvation avoids these by approximating bulk solvent effects without adding solvent degrees of freedom \cite{van2019implicit_neb_1, hossain2020implicit_neb_2}.

\paragraph{\markdiff{Solvent-Aware Reaction Pathway Prediction}}

\markdiff{A key application of MLIP is the construction of a minimum-energy pathway (MEP). In realistic settings, however, MEP calculations must often account for solvent effects, which remain challenging for existing MLIPs to capture directly. A common workaround is to include explicit solvent molecules using the existing MLIPs, but this is often highly sensitive to solvent configuration, hence requiring sampling methods for reliable predictions such as umbrella sampling. \ourmodel{}, on the other hand, can encode environmental effects, which provide a more practical solution for solvent-aware MEP construction. We discuss this further in Section \ref{sec:env_effects}.}

% This reliable generalization of \foundationmodel{} to unseen or underrepresented chemistry enables real-world applications such as minimum-energy pathway (MEP) construction. In realistic settings, however, MEP calculations must often account for solvent effects, which remain challenging for existing MLIPs to capture directly. A common workaround is to include explicit solvent molecules using the existing MLIPs such as UMA, but this substantially increases system size and makes computational pathway methods such as NEB highly sensitive to solvent configuration. By encoding environmental effects, including implicit solvation, \foundationmodel{} provides an even more practical solution for solvent-aware MEP construction.

%\ourmodel{} can account for implicit solvation, providing a tractable way to account for solvent effects for reactive systems. 
\markdiff{To build a practical solvent-aware reactive MLIP, we create the \tlxsolv{} dataset with reaction geometries evaluated under four solvent conditions (vacuum, water, methanol, and toluene), and fine-tune the pretrained \foundationmodel{} model on \tlxsolv{} to leverage learned representations. Further details on the dataset are provided in Section \ref{sec:datasets_and_trainings}.}

\begin{figure}[!ht]
    \centering
    \includegraphics[width=\linewidth]{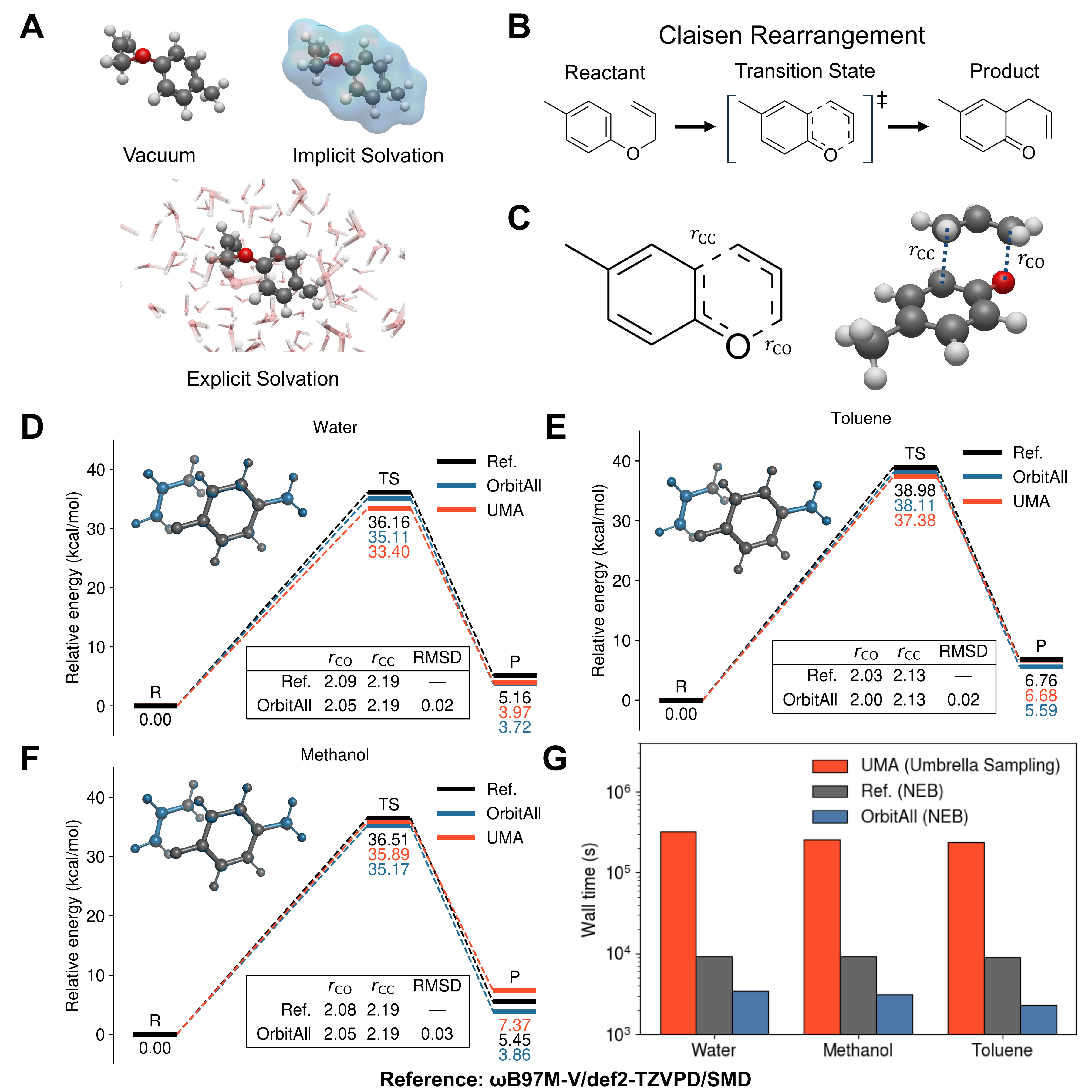}
    \caption{\markdiff{\textbf{Transition-state (TS) search under different solvents.} \textit{(A)} Comparison of implicit and explicit solvation: implicit solvation represents averaged solvent effects as a field, whereas explicit solvation treats individual solvent molecules. \textit{(B)} Claisen rearrangement of allyl-\textit{p}-tolyl ether (APTE). \textit{(C)} Schematic and 3D structure of the transition state, with the forming and breaking bond lengths ($r_\text{CO}$ and $r_\text{CC}$) labeled. \textit{(D--F)} TS search results for APTE in different solvents. ``Ref.'' denotes reference calculations at the $\omega$B97M-V/def2-TZVPD/SMD level of theory. Reaction barriers (TS) and reaction energies (P) from each method are labeled below the corresponding states. Ref. and \ourmodel{} results are obtained using NEB, whereas UMA results are obtained using TS-informed umbrella sampling. Overlaid TS structures are shown with Ref. in gray and \ourmodel{} in blue; $r_\text{CO}$, $r_\text{CC}$, and RMSD relative to Ref. are reported in \AA{} in the inset tables. \textit{(G)} Wall-time comparison across methods. UMA wall time is measured for TS-informed umbrella sampling as an explicit solvation baseline, and Ref. with GPU4PySCF and \ourmodel{} wall times are measured for NEB with implicit solvation.}}
    \label{fig:neb}
\end{figure}

\markdiff{We use the Claisen rearrangement of allyl-\textit{p}-tolyl ether as an example to predict the MEP with NEB using \ourmodel{} and the reference method ($\omega$B97M-V/def2-TZVPD/SMD). The reaction schematic and a representative transition-state (TS) structure are shown in Figures \ref{fig:neb}(B) and \ref{fig:neb}(C), respectively. For both \ourmodel{} and the reference, we perform NEB simulations under different implicit solvents; experimental details are provided in Section \ref{sec:method_experiments}. We also compare to UMA (UMA-s-1p2) \cite{wood2025uma} to provide an explicit solvation baseline with umbrella sampling. Details of the experiments are provided in Section \ref{sec:method_experiments}.}

\markdiff{The fine-tuned \ourmodel{} shows strong performance in predicting the TS energetics and structures of the Claisen rearrangement across all solvent conditions (Figures \ref{fig:neb}(D--F)). The predicted reaction barriers ($\Delta E^\ddagger$) and reaction energies ($\Delta_r E$) closely match the reference values in all solvents. For the TS structures, \ourmodel{} accurately predicts the bond-breaking/forming distances, $r_\text{CO}$ and $r_\text{CC}$, with overall RMSDs of only 0.02--0.03 \AA{}. Notably, \ourmodel{} captures subtle solvent-dependent structural trends, including the longer TS bond lengths in water and methanol than in toluene. It also reproduces the corresponding energetic trend, predicting a higher reaction barrier in toluene than in water and methanol. Compared with the explicit-solvent umbrella-sampling baseline using UMA, \ourmodel{} achieves reference-level accuracy at approximately two orders of magnitude lower computational cost, providing a scalable approach for TS search under implicit solvation.}

\markdiff{Additionally, \ourmodel{} accurately predicts the Diels-Alder reaction between cyclopentadiene and methyl vinyl ketone (\SI{}Figure \ref{fig:app:neb_da}). Remarkably, it captures the higher reaction barrier in toluene than in vacuum, as well as the lower barriers in water and methanol than in vacuum. Furthermore, the generalization experiment shows that \ourmodel{} can generalize to solvents similar to those in the training dataset (\SI{}Figure \ref{fig:app:neb_other_solvs}), highlighting the need for datasets with broader solvent coverage.}

% Furthermore, the generalization experiment shows that \ourmodel{} can generalize to solvents similar to those in the training dataset (\SI{}Figure \ref{fig:app:neb_other_solvs}), highlighting the need for datasets with broader solvent coverage. The model performs well for benzene and isopropanol: benzene is similar to toluene as a nonpolar solvent, whereas isopropanol is similar to water and methanol as a polar hydrogen-bond-donating solvent. In contrast, the model fails to generalize to acetone and acetonitrile, which are polar aprotic solvents not represented in the dataset. This limitation likely arises from the simplified generalized Born with finite epsilon (GBE) implicit solvation method used with g-xTB \cite{froitzheim2025gxtb}. A more sophisticated implicit solvation method and a more diverse solvent dataset may improve generalization. Nonetheless, these results provide encouraging evidence that the model can generalize to unseen solvents that are chemically similar to those included in the training dataset, while also highlighting the need for broader solvent coverage.

\subsection{Molecular Systems with Different Spins and Charges}
\label{sec:qm9star_spin_charge}

To accurately predict properties of charged and/or open-shell species, spin and charge information must be incorporated, quantities often missing from the input data structure of geometry-based GNN models \cite{batzner2022nequip, schutt2018schnet, schutt2021painn, batatia2022mace, liao2023equiformerv2, gasteiger2022dimenetpp}. 
These quantities are inherently included in the orbital features of \ourmodel{}, providing physical information such as electronic distribution and intensity of spin polarization.

We compare \ourmodel{}'s performance to seven geometric GNNs: SpookyNet \cite{unke_spookynet_2021}, SchNet \cite{schutt2018schnet}, PaiNN \cite{schutt2021painn}, DimeNet++ \cite{gasteiger2022dimenetpp}, SphereNet \cite{liu2022spherenet}, NequIP \cite{batzner2022nequip}, and EquiformerV2 \cite{liao2023equiformerv2}.
To incorporate charge and spin into different geometric GNN models, we augment the node representations of these models with additional channels for spin and charge embeddings, as introduced in \cite{unke_spookynet_2021}.
Modified models are labeled “-SC” to indicate inclusion of spin and charge. Models trained on delta-labels are marked ``($\Delta$)''.

\paragraph{Evaluation on Varying Spins and Charges}

We use QM9star \cite{tang2024qm9star}, a QM9-based dataset \cite{ramakri_qm9_2014} with neutral open-shell doublet molecules (radicals), closed-shell charged molecules (anions and cations), and neutral closed-shell molecules (neutrals). Each molecular property is calculated at the B3LYP-D3(BJ)/6-311+G(d,p) level of theory. We evaluate \ourmodel{}'s performance on two tasks: (1) the total energy ($E_\text{tot}$) of molecules with varying spins and charges; and (2) the HOMO and LUMO energy levels of radical species for both spins, $\alpha$ and $\beta$ ($E^\alpha_\text{HOMO}$, $E^\alpha_\text{LUMO}$, $E^\beta_\text{HOMO}$, $E^\beta_\text{LUMO}$).

\begin{figure}[ht]
    \centering
    \includegraphics[width=1\textwidth]{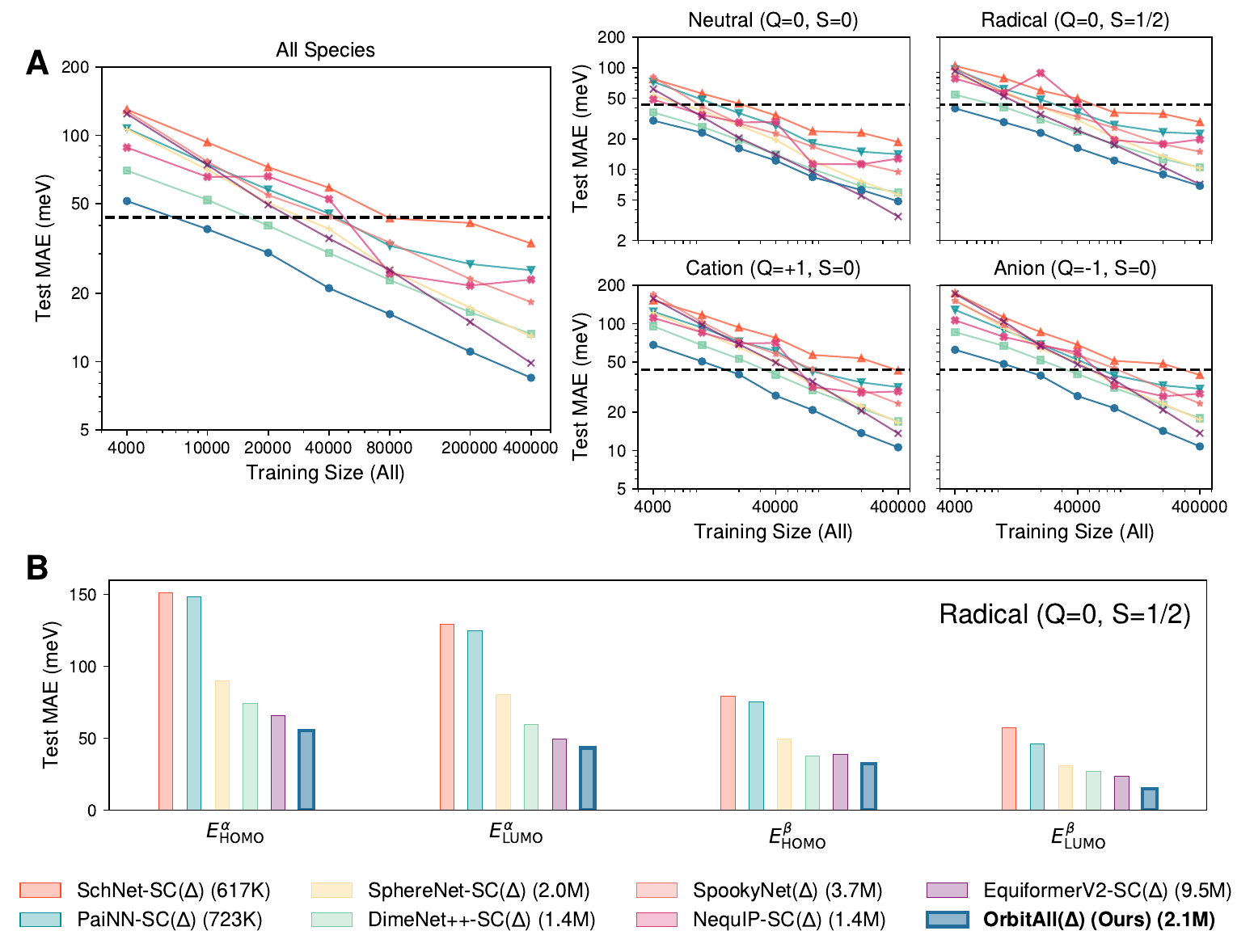}
    \caption{\textbf{The QM9star (B3LYP-D3(BJ)/6-311+G(d,p)) benchmarks with delta-learning.} The numbers of learnable parameters are indicated inside brackets next to model names in the legend. \textit{(A)} Learning curves of different models for all species and for each category (neutral, radical, cation, anion). Each category's learning curve of different models for the total energy $E_\text{tot}$ is plotted. For training size $N$, the training set contains $N/4$ neutrals, radicals, cations, and anions each. The dotted black lines indicate chemical accuracy (1 kcal/mol $\approx$ 43.4 meV). \textit{(B)} HOMO and LUMO energy levels prediction benchmark of radical species for up-spin ($\alpha$) and down-spin ($\beta$). The test mean absolute errors (MAEs) of different models trained using 100K radicals are presented for each property ($E^\alpha_\text{HOMO}$, $E^\alpha_\text{LUMO}$, $E^\beta_\text{HOMO}$, $E^\beta_\text{LUMO}$).}
    \label{fig:qm9star_learning_curves}
\end{figure}

For the total energy $E_\text{tot}$ prediction, we train all models with the delta-learning strategy. This enables assessing the advantage of orbital learning separately, by comparing against purely geometry-based models.
We evaluate each model using training data of varying sizes. All training and validation datasets are composed of equally distributed species (neutral, radical, cation, anion). The largest training set consists of 400K data (100K for each species), 20K validation data (5K for each species), and $\sim$75K test data ($\sim$15K neutrals, 20K for other species). We keep the species ratio the same in all smaller training and validation sets.

As shown in Figure \ref{fig:qm9star_learning_curves}(A), \ourmodel{} outperforms the other models consistently on the combined test set for all training sets of different sizes. 
Additionally, \ourmodel{} outperforms all other models for each separate species across all training sets of different sizes except for the neutral subset in training sizes above 200K. 
\ourmodel{} is especially valuable in predicting non-neutral species (radical, anion, cation), with smaller differences between the $E_\text{tot}$ mean absolute errors (MAEs) of neutral and non-neutral species (\SI{}Table \ref{tab:app:qm9star_summary_table}). 

Furthermore, delta-learning enhances generalization as well. For QM9star, we observe that training with the direct-learning strategy leads to consistently higher errors for cations than anions across all models (\SI{}Table \ref{tab:app:qm9star_summary_table})--a result that is not observed when training with the delta-learning strategy. One explanation for this is the distribution shifts of the labels, where the delta-labels are easier to predict, especially for charged species (\SI{}Figure \ref{fig:app:qm9star_direct_delta_label_dist}). 

We also use \ourmodel{} to predict frontier molecular orbital (FMO) energy levels for the QM9star dataset.
For radical species, four FMO energy levels exist: $E^\alpha_\text{HOMO}$, $E^\alpha_\text{LUMO}$, $E^\beta_\text{HOMO}$, and $E^\beta_\text{LUMO}$. We evaluate and compare the performance of \ourmodel{} in predicting all these levels with delta-learning. 
Since unpaired electrons cause spin polarization and split spatial orbital energy levels by spin, predicting FMO levels for radicals is expected to be more challenging than for closed-shell species.
As shown in \ref{fig:qm9star_learning_curves}(B), similar to $E_\text{tot}$, \ourmodel{} outperforms other competing models across the different FMO levels, demonstrating the effectiveness of spin-polarized orbital features for predicting such quantities. 

We present additional benchmarks and comparisons for predicting the singlet and triplet energies and their gaps of carbene molecules (the QMSpin dataset) in \SI{}\appendixsection{} \ref{app:qmspin_result}. \ourmodel{} achieves the lowest MAEs in nearly all categories, with singlet/triplet errors and vertical/adiabatic spin-gap MAEs below chemical accuracy. \ourmodel{} uses a unified joint model, which likely improves robustness across spin multiplicities.

\paragraph{Cost-Accuracy Analysis}

To employ delta-learning, QC calculations are needed, which are involved for \ourmodel{} during feature generation. To develop a description of the cost-accuracy tradeoff, we compare \ourmodel{} with direct-learned geometric GNNs.  

\begin{figure}[ht]
    \centering
    \includegraphics[width=1\linewidth]{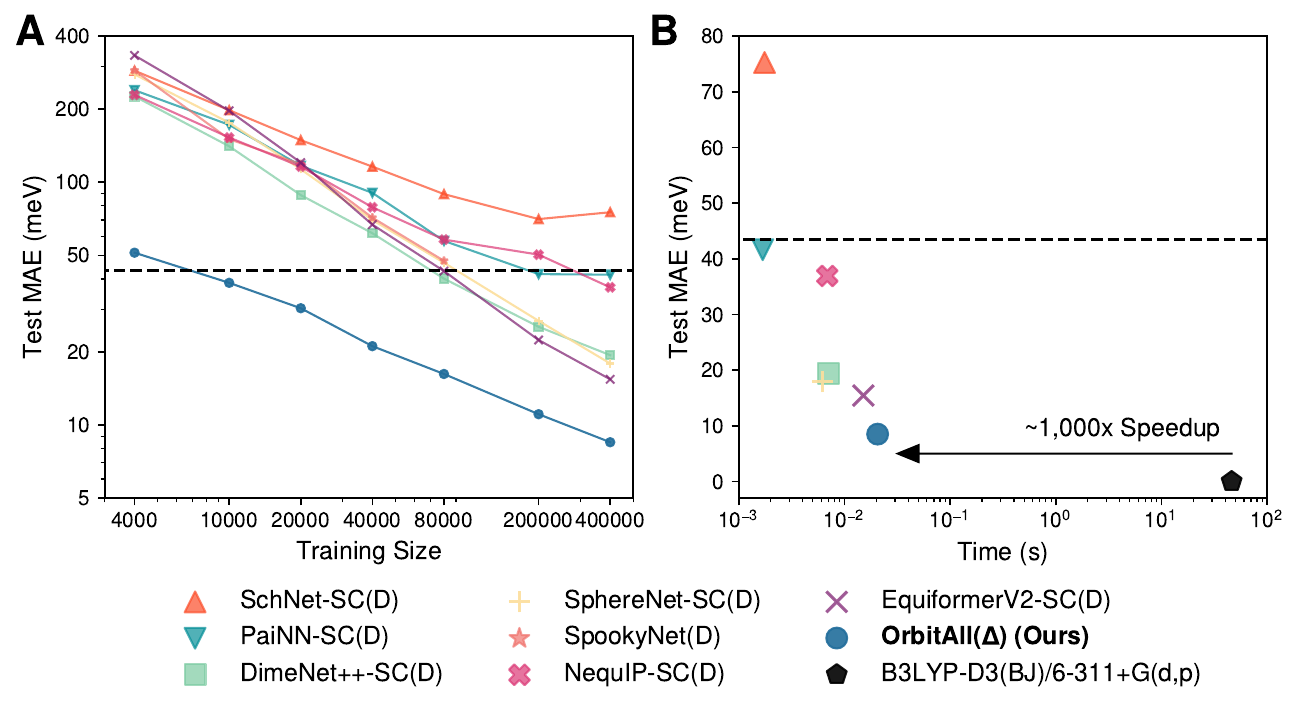}
    \caption{\textbf{Cost, accuracy, and data-efficiency comparisons.} ``(D)'' indicates direct-learning and ``($\Delta$)'' indicates delta-learning \cite{rama_delta_learning_2015}. Black dotted lines represent chemical accuracy (1 kcal/mol $\approx$ 43.4 meV). \textit{(A)} Data-efficiency comparison using learning curves of different models. \ourmodel{} offers greater data-efficiency, requiring $\sim$10 times less data than the next best model (DimeNet++-SC(D)) to achieve chemical accuracy on the QM9star dataset. \textit{(B)} Cost-accuracy comparisons between different methods on QM9star. \ourmodel{} achieves $\sim$1,000 times speedup compared to the QM, B3LYP-D3(BJ)/6-311+G(d,p). All models are trained using the 400K training set of QM9star.}
    \label{fig:cost_vs_accuracy}
\end{figure}

Delta-learning typically leads to a reduction of errors with a certain offset to the learning curves. This brings a significant advantage in data-efficiency compared to directly learning the targets, as shown in Figure \ref{fig:cost_vs_accuracy}(A). By interpolation, we find that \ourmodel{} requires $\sim$7K training data to achieve chemical accuracy, whereas the next-best model (DimeNet++-SC(D)) requires $\sim$70K training data to achieve chemical accuracy. Furthermore, shown in Figure \ref{fig:cost_vs_accuracy}(B), \ourmodel{} achieves about 1,000-fold speedup compared to the original DFT method while still providing a near-chemical accuracy MAE in the polypeptide dataset. This means that \ourmodel{} can accelerate simulations by several orders of magnitude while retaining accuracy. Although orbital learning introduces additional computational costs, \ourmodel{} achieves comparable acceleration while retaining its key advantages, such as data-efficiency.

\paragraph{Transferability to Much Larger Molecules than in the Training Set}
\label{sec:extrapolation-polypeptides}

A key advantage of predictive systems is their applicability to large molecular systems where traditional quantum mechanical simulations become impractical. However, most AI models exhibit degraded performance when applied to molecules outside their training set.
To assess the ability of models to extrapolate across molecular systems of such sizes, we created the \textit{polypeptide dataset}, consisting of 151 data points covering neutral, radical, cation, and anion species. The models trained on the 400K QM9star training set (Section \ref{sec:qm9star_spin_charge}) are used to predict total energies ($E_\text{tot}$) of the polypeptides, computed at the B3LYP-D3(BJ)/6-311+G(d,p) level of theory (the same level of theory as QM9star).

\begin{figure}[!ht]
    \centering
    \includegraphics[width=\linewidth]{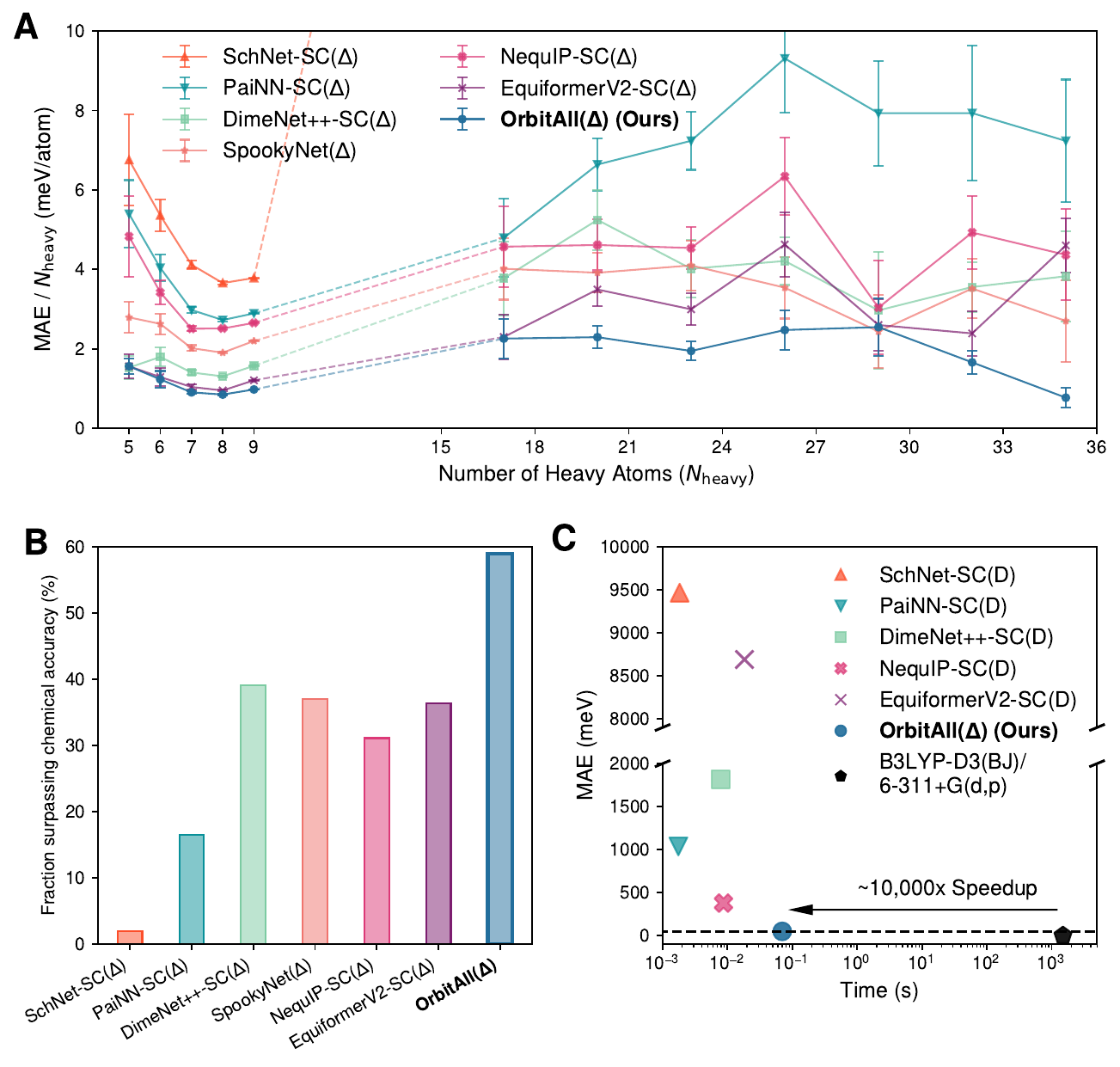}
    \caption{\textbf{Size-wise extrapolation performance of different models evaluated using the polypeptide dataset for the total energy $E_\text{tot}$ prediction on 151 polypeptides.} All models are trained using the 400K training set of QM9star. ``(D)'' indicates direct-learning and ``($\Delta$)'' indicates delta-learning. \textit{(A)} Size-wise extrapolation capabilities. The MAEs normalized by the number of heavy atoms (C, N, O) is shown as a function of heavy atom count. The five points on the left side are the normalized MAEs for the QM9star test set. The remaining points on the right side are the normalized MAEs for the polypeptide dataset inference, which are grouped into bins of width 3 (e.g., 16–18, 19–21, …) based on the number of heavy atoms. The MAE and standard error of the mean (SEM) are reported across systems within the bin/point. \textit{(B)} Proportion of predictions of $E_\text{tot}$ surpassed chemical accuracy (1 kcal/mol $\approx$ 43.4 meV). \ourmodel{} records $\sim$1.5 times more predictions within chemical accuracy than the next best model. \textit{(C)} Cost-accuracy comparisons between different models on the polypeptide dataset. The black dotted line indicates chemical accuracy. \ourmodel{} achieves near-chemical accuracy MAE (48.6 meV), with $\sim$10,000 times speedup compared to the original method, B3LYP-D3(BJ)/6-311+G(d,p).}
    \label{fig:polypeptides_err_curve}
\end{figure}

Figure \ref{fig:polypeptides_err_curve}(A) shows MAEs normalized by the number of heavy atoms ($N_\text{heavy}$) across $N_\text{heavy}$ ranges for both QM9star and the polypeptide dataset. \ourmodel{} shows the smallest increase in normalized error, maintaining consistently low MAEs and demonstrating superior extrapolation to larger molecules. As a result, \ourmodel{} yields the highest fraction of $E_\text{tot}$ predictions within chemical accuracy (Figure~\ref{fig:polypeptides_err_curve}(B)). Figure~\ref{fig:polypeptides_err_curve}(C) compares the cost and accuracy of different models on the polypeptides; as in Figure~\ref{fig:cost_vs_accuracy}(B), only models trained with the direct-learning strategy (excluding \ourmodel{}) are included. Owing to the larger size and electron count of polypeptides compared to QM9star molecules, the speedup is even greater than that in Figure~\ref{fig:cost_vs_accuracy}(B).
Solely recording MAE with near-chemical accuracy (48.6 meV) on average, \ourmodel{} serves as a method with balanced speed and accuracy without sacrificing either.

\subsection{Molecular Systems Under Various Environmental Effects}
\label{sec:env_effects}

% Predicting the response of molecular systems to external perturbations from the environment is necessary for calculating important molecular properties. Models typically use input variables that represent the environment, such as the dielectric constant for implicit solvents or the electric field vector, to account for the environmental effects \cite{gastegger2021fieldschnet, ward2021solvent_mpnn, falletta2025electricresponse}.

Predicting the response of molecular systems to external perturbations from the environment is necessary for calculating important molecular properties. Without explicit embedding layers (e.g., Ref. \cite{gastegger2021fieldschnet, zhang2026consolv}), \ourmodel{} naturally incorporates environmental effects into a common representation space, since the orbital features obtained by the SCF procedure are directly perturbed by the environment. 
Environmental effects, such as external electric fields and solvation via implicit treatment, can be applied within the semi-empirical QM simulation (e.g., GFN$n$-xTB, g-xTB).% For example, we can use the conductor-like polarizable continuum model (CPCM) for implicit solvent, where the converged electronic structure is altered by different polarities of solvents \cite{barone1998cpcm1, takano2005cpcm2, alibakhshi2022orbitalfeature_perturbed}.

\paragraph{Molecular Systems in Solvents}

We trained a single \ourmodel{} model to predict the total energies $E_\text{tot}$ of molecules in the four solvent environments: vacuum, water, tetrahydrofuran (THF), and toluene. Orbital features were generated using the ALPB method \cite{sigalov2006alpb,ehlert2021alpb2}, chosen after ablation studies with other implicit solvation methods (\SI{}Table \ref{tab:app:hessian_qm9_diff_solvents})

\begin{figure}[ht]
    \centering
    \includegraphics[width=\linewidth]{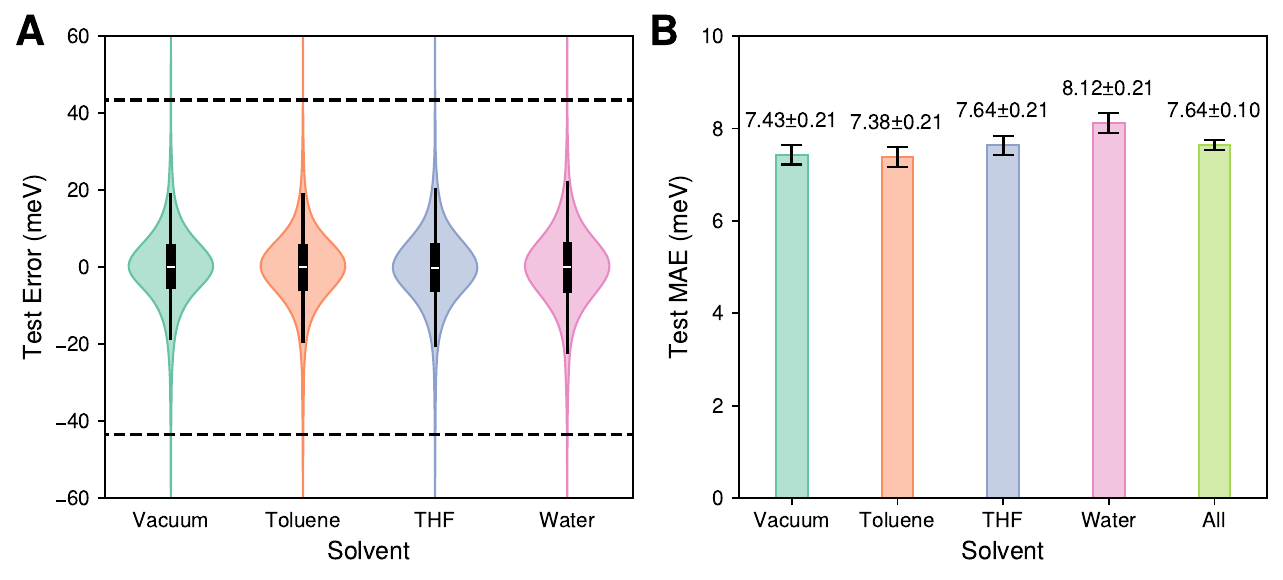}
    \caption{\textbf{Evaluations of \ourmodel{} on Hessian QM9 ($\omega$B97x/6-31G*).} \ourmodel{} is trained using 128K training data (32K of each solvent). \textit{(A)} Violin plots of error distributions for different solvents. The dotted black lines indicate chemical accuracy ($\pm$1 kcal/mol $\approx$ 43.4 meV). \textit{(B)} Mean absolute errors (MAEs) for molecules in different solvents and the average (``All''). The error bars represent the standard errors of mean (SEMs).}
    \label{fig:hessian_qm9_result}
\end{figure}

As shown in Figure \ref{fig:hessian_qm9_result}(A), the error distributions for different solvents are almost identical, demonstrating the robust performance of \ourmodel{} for different solvents. Compared to baselines provided in \cite{williams2025hessianqm9}, \ourmodel{} records about 3-4 times smaller MAE across different solvents. Additionally, the MAEs for different solvents are very close while also being significantly smaller than chemical accuracy (Figure \ref{fig:hessian_qm9_result}(B)).

An example of this perturbation is illustrated in \SI{}Figure \ref{fig:app:hessian_qm9_solvent_comparison}. In addition, we conducted another set of experiments using uniform external electric fields as an environmental effect. The results are presented in \SI{}\appendixsection{} \ref{app:qm9star_in_random_efield}.

\section{Discussion}
\label{sec:discussion}

We present \ourmodel{}, a physics-informed end-to-end SE(3)-equivariant deep learning framework designed to process all molecular systems by simultaneously accounting for varying charges, spin states, and environmental effects. We demonstrate the accuracy and robustness of \ourmodel{} by evaluating its performance using different quantum chemistry datasets of all molecular systems. \ourmodel{} consistently outperforms other machine learning approaches, mainly due to its ability to integrate physically-grounded information across these different conditions--conditions that other models often treat separately, if at all. \ourmodel{} incorporates physical effects through SCF-derived orbital features, providing a practical route toward more unified quantum-chemical learning frameworks.

% Notable examples with such treatment include SpookyNet \cite{unke_spookynet_2021} and TensorNet \cite{simeon2025tensornet_spincharge}, which introduce mechanisms to handle spin and charge variations but do not generalize to environmental perturbations. Incorporating environmental effects remains particularly challenging, as their diversity and complexity increase input dimensionality and often lead to optimization difficulties. In contrast, \ourmodel{} naturally captures environmental effects by embedding highly physics-informed information into molecular representations. Furthermore, as demonstrated in our cost-accuracy and size-wise extrapolation analyses, such extensions typically require significantly more data to achieve chemical accuracy and still exhibit limited transferability. \ourmodel{} addresses this challenge by incorporating these effects through SCF-derived orbital features, providing a practical route toward more unified quantum-chemical learning frameworks.

Incorporating environmental effects remains particularly challenging, as their diversity and complexity increase input dimensionality and often lead to optimization difficulties. \markdiff{In contrast, \ourmodel{} naturally captures environmental effects by embedding physics-informed information into molecular representations, as shown by the accurate predictions of reaction energetics and TS structures for reactions under implicit solvents. These capabilities could accelerate realistic-condition reaction prediction without requiring expensive DFT calculations or MLIP models in which solvent effects are added post hoc.}

% Furthermore, as demonstrated in our cost-accuracy and size-wise extrapolation analyses, molecular condition embedding extensions typically require significantly more data to achieve chemical accuracy and still exhibit limited transferability. \ourmodel{} addresses this challenge by incorporating these effects through SCF-derived orbital features, providing a practical route toward more unified quantum-chemical learning frameworks.

% \ourmodel{} enables property predictions that are $10^3-10^4$ faster compared to DFT without sacrificing accuracy. This high accuracy is achieved with strong data efficiency, which is essential when working with costly quantum chemical or experimental data. This data efficiency suggests that orbital-feature-based models such as \ourmodel{} may be useful for future data-scarce applications, such as MOF band-gap prediction \cite{rosen2021qmof1, rosen2022qmof2} or experimentally labeled molecular properties.

\ourmodel{} enables property prediction at DFT-level accuracy while achieving a $10^3$–$10^4$-fold speedup over DFT.
The accuracy on par with high-level QM simulations is achieved with strong data efficiency, which is essential when working with costly quantum chemical data. This data efficiency suggests that orbital-feature-based models such as \ourmodel{} may be useful for future data-scarce applications, such as experimentally labeled molecular properties.

% Predictions on chemically diverse and significantly large species demonstrate that \ourmodel{} generalizes effectively across a wide range of chemical conditions. This enables future studies toward building multi-purpose foundational models using diverse datasets \cite{ganscha2025qcml_dataset, levine2025openmolecules2025omol25} that span elements, spin states, charges, and environments. Foundational models built upon \ourmodel{}'s universal representation can robustly handle complex scenarios—such as molecules under external electric fields—where current foundational models like EquiformerV2 \cite{liao2023equiformerv2} and UMA \cite{wood2025uma} could struggle.

One limitation of \ourmodel{} is its reliance on the convergence of the underlying SCF method in the low-level semi-empirical QM calculations. Since the neural network depends on QMMs generated from these simulations, failure of the SCF procedure prevents feature generation and, consequently, prediction. A possible future study to mitigate this limitation is to use more robust alternatives, such as non-self-consistent field methods like GFN0-xTB \cite{pracht2019gfn0xtb}.

\section{Methods}
\label{sec:methods}

\subsection{The \ourmodel{} Framework}

\paragraph{Unrestricted Open-shell Features}

In QM calculations, electron spin can be treated using either restricted or unrestricted methods. Restricted calculations (e.g., restricted Hartree-Fock, RHF) constrain the spin orbitals within a spatial orbital to have the same energy level whereas unrestricted methods (e.g., unrestricted Hartree-Fock, UHF) account for spin polarization arising from unequal numbers of spin-up and spin-down electrons. Depending on the system, one may choose between restricted and unrestricted open-shell treatments. In this work, we adopt an unrestricted formulation to generate orbital features, enabling the model to represent a wider range of molecular systems, including open-shell and spin-polarized species \cite{szabo_quantum_1989}.

For unrestricted open-shell systems, the Pople-Nesbet equations, equivalent to the Roothaan-Hall equation for restricted Hartree-Fock, can be expressed as:

\begin{equation}\label{eqn:pople_nesbet}
    \textbf{F}^\alpha\textbf{C}^\alpha=\textbf{S}\textbf{C}^\alpha\mathbf{\varepsilon}^\alpha, \qquad \textbf{F}^\beta\textbf{C}^\beta=\textbf{S}\textbf{C}^\beta\mathbf{\varepsilon}^\beta,
\end{equation}

\noindent
where $\alpha$ denotes up-spin, $\beta$ denotes down-spin, $\textbf{F}^\alpha$ and $\textbf{F}^\beta$ are the Fock matrices, $\textbf{C}^\alpha$ and $\textbf{C}^\beta$ are orbital coefficients matrices, $\textbf{S}$ is the overlap matrix, and $\mathbf{\varepsilon}^\alpha$ and $\mathbf{\varepsilon}^\beta$ are diagonal orbital energy matrices. Also, the density matrices for different spins are given by:

\begin{equation}
    P^\alpha_{\mu\nu}=\sum^{N_\alpha}_a C^\alpha_{\mu a}(C^{\alpha}_{\nu a})^*, \qquad
    P^\beta_{\mu\nu}=\sum^{N_\beta}_a C^\beta_{\mu a}(C^{\beta}_{\nu a})^*,
\end{equation}

\noindent
where $P^\alpha_{\mu\nu}$ and $P^\beta_{\mu\nu}$ are elements of the density matrices $\textbf{P}^\alpha$ and $\textbf{P}^\beta$, respectively, for orbital bases $\mu$ and $\nu$. $N_\alpha$ and $N_\beta$ are the numbers of up-spin electrons and down-spin electrons, respectively. $C^\alpha_{\mu\nu}$ and $C^\beta_{\mu\nu}$ are elements of the orbital coefficient matrices $\textbf{C}^\alpha$ and $\textbf{C}^\beta$.

The unrestricted Hartree-Fock (UHF) formulation generalizes the standard Hartree-Fock approach to handle open-shell systems. Specifically, for closed-shell molecules, the resulting UHF matrices reduce to those of the restricted Hartree-Fock (RHF) method in a way that:

\begin{equation}
    \textbf{F}^\alpha=\textbf{F}^\beta=\textbf{F},
\end{equation}

\begin{equation}
\textbf{P}^\alpha=\textbf{P}^\beta=\frac{1}{2}\textbf{P},
\end{equation}

\noindent
where $\textbf{P}$ is the density matrix for restricted closed-shell Hartree-Fock \cite{szabo_quantum_1989}. The density matrix $\textbf{P}$ for restricted closed-shell Hartree-Fock is then equivalent to the total density matrix $\textbf{P}^T=\textbf{P}^\alpha+\textbf{P}^\beta$ for unrestricted Hartree-Fock. This equivalence enables a unified input representation that can accommodate both open- and closed-shell systems. It is important to note that this assumes the molecule is a closed-shell singlet, not an open-shell singlet.

The QMMs are SE(3)-equivariant as shown in (\ref{eqn:qcm_block_rotation}). The set of QMMs for \ourmodel{}, $\textbf{T}=(\textbf{F}^\alpha,\textbf{F}^\beta,\textbf{P}^\alpha,\textbf{P}^\beta,\textbf{S},\textbf{H}_\text{core})$, is used as inputs to the equivariant graph neural network backbone, UNiTE \cite{qiao_orbnet_equi_2022}. These features vary with molecular conditions such as spin, charge, and environmental effects. For example, the electron densities for up-spin and down-spin at position $\textbf{r}$ is given by:

\begin{equation}
    \rho^\alpha(\textbf{r})=\sum_{\mu}\sum_\nu (\textbf{P}^\alpha)^{\mu,\nu}\Phi^\mu(\textbf{r})(\Phi^{\nu}(\textbf{r}))^*,
\end{equation}

\begin{equation}
    \rho^\beta(\textbf{r})=\sum_{\mu}\sum_\nu (\textbf{P}^\beta)^{\mu,\nu}\Phi^\mu(\textbf{r})(\Phi^\nu(\textbf{r}))^*,
\end{equation}

\noindent
where $\Phi^\mu(\textbf{r})$ and $\Phi^\nu(\textbf{r})$ are the atomic orbital bases. By definition, they satisfy

\begin{equation}
    \int d\textbf{r}\,(\rho^\alpha(\textbf{r})+\rho^\beta(\textbf{r}))=N_\text{elec},
\end{equation}

\begin{equation}
    \int d\textbf{r}\,(\rho^\alpha(\textbf{r})-\rho^\beta(\textbf{r}))=2S,
\end{equation}

\noindent
where $N_\text{elec}$ denotes the number of electrons, which depends on the system’s charge. As a result, the density matrix inherently encodes both spin and charge information. Environmental effects influence the density matrix more subtly by altering the converged mean-field compared to that of an isolated system. An example of such a perturbation is shown in \SI{}Figure \ref{fig:app:hessian_qm9_solvent_comparison}.

Any SCF method solving the Roothaan-Hall or Pople-Nesbet equations can be used in this framework. We use spGFN1-xTB \cite{neugebauer_spgfnxtb_2023} for the experiments in Sections \ref{sec:qm9star_spin_charge} and \ref{sec:env_effects}, and g-xTB for the experiments in Section \ref{sec:foundation_model} \cite{froitzheim2025gxtb}, to efficiently generate orbital features with unrestricted Hartree-Fock. 

% Other relatively inexpensive methods, such as spGFN2-xTB \cite{bannwarth2019gfn2} or recently released g-xTB \cite{froitzheim2025gxtb}, can be used to possibly improve the accuracy of the predictions.

\paragraph{Equivariant Graph Neural Network}

To build a data-efficient model that robustly predicts SE(3) (rotations and translations) equivariant and/or invariant molecular properties such as forces and dipole moments, we construct a GNN, based on the E(3) equivariant UNiTE framework \cite{qiao_orbnet_equi_2022}. 
The inherent SE(3) equivariance of the QMMs is maintained through our GNN, which results in an end-to-end SE(3) equivariant framework. The GNN satisfies the necessary rotational and translational symmetries:

\begin{equation}\label{eqn:equivariance}
    \mathcal{R}\cdot\mathcal{F}(\textbf{T})=\mathcal{F}(\mathcal{R}\cdot\textbf{T}),
\end{equation}

\noindent
where $\mathcal{F}$ characterizes the parameters of the GNN, $\mathcal{R}$ is an arbitrary roto-translational operation, and $\mathcal{R}\,\cdot$ represents applying the roto-translational transformation $\mathcal{R}$.
The geometric GNN learns to map a set of QMMs of any molecular system, $\textbf{T}$, to an atomic or molecular property, by minimizing the objective:

\begin{equation}\label{eqn:loss_min}
    \min_{\mathcal{F}}\mathcal{L}(y,\hat{y}),
\end{equation}

\noindent
where $\mathcal{L}$ is the loss function specific to the predicted property, $y$ is the target property, either generated by simulation or estimated by experiments, and $\hat{y}=\mathcal{F}(\textbf{T})$.

\paragraph{Diagonal Reduction and Embedding}

All QMMs have the same dimensions of $(N_\text{AO},N_\text{AO})$, where $N_\text{AO}$ is the number of atomic orbitals in the system. Each row and column corresponds to an atomic orbital. The first $N_\text{AO}^A$ entries of rows and columns correspond to the atomic orbitals of atom $A$, the next $N_\text{AO}^B$ to those of atom $B$, and so on. As a result, the block-diagonal regions of the QMMs capture intra-atomic interactions, while the off-diagonal blocks encode inter-atomic interactions. These inter-atomic blocks are later used to construct messages in the model, as illustrated in Figure \ref{fig:overview}(C).

From the atomic orbital basis QMMs, we construct an atom-based representation via the diagonal reduction module. This module embeds the block-diagonals of QMM $\textbf{O}$ to the reduced embeddings $\textbf{h}^{O}_A$, as follows:

\begin{equation}
    \textbf{h}^O_{A,nlpm}=\begin{cases}
        \sum_{\mu,\nu} (\textbf{O})^{\mu,\nu}_{AA} (\tilde{\textbf{Q}})^{\mu,\nu}_{A,nlm}, & p=+1,\\
        0, & p=-1,
    \end{cases}
\end{equation}

\noindent
where $p$ is the parity and $\tilde{\textbf{Q}}$ is an on-site three-index overlap integrals, defined by:

\begin{equation}
    (\tilde{\textbf{Q}})^{\mu,\nu}_{A,nlm}=\int_{\mathbb{R}^3}d\textbf{r}\,(\Phi^\mu_A(\textbf{r}))^*\Phi^\nu_A(\textbf{r})\tilde{\Phi}^{n,l,m}_A(\textbf{r}),
\end{equation}

\noindent
where $\tilde{\Phi}^{n,l,m}_A(\textbf{r})$ is an auxiliary Gaussian-type basis as defined in \cite{qiao_orbnet_equi_2022}. Note that $\tilde{\textbf{Q}}$ is proportional to the Clebsch-Gordan coefficients, ensuring the overall process is equivariant to SO(3).

Since g-xTB and spGFN1-xTB employ different basis sets, we compute and tabulate the three-index overlap integrals separately for each method. In both approaches, atomic orbitals are constructed from primitive Gaussian-type orbitals (GTOs). However, while spGFN1-xTB uses fixed contraction coefficients, g-xTB employs environment-dependent (i.e., charge-dependent) contraction coefficients~\cite{froitzheim2025gxtb}. Specifically, for an atom $A_i$ of element type $A$, an AO basis function $\Phi_{A_i}^{\mu}$ is expressed as a contraction over primitive GTOs:

\begin{equation}
\Phi^\mu_{A_i}(\mathbf r)
=\sum_{\lambda}
c^{\mu\lambda}_{A_i}\left(q^\mathrm{eff}_{A_i}\right)\,
\chi^\lambda_{A}(\mathbf r;\zeta^\lambda_A),
\end{equation}

\noindent
where $\lambda$ indexes the primitive GTOs associated with element $A$, $\chi^\lambda_A(\mathbf r;\zeta^\lambda_A)$ denotes a primitive GTO with exponent $\zeta^\lambda_A$, and $c^{\mu\lambda}_{A_i}(q^\mathrm{eff}_{A_i})$ are contraction coefficients that depend explicitly on the effective atomic charge $q^\mathrm{eff}_{A_i}$.

Because the contraction coefficients vary with the chemical environment through $q^\mathrm{eff}_{A_i}$, the on-site three-index overlap integrals cannot be tabulated directly in the contracted AO basis. Instead, we tabulate the corresponding quantities in the primitive GTO basis,

\begin{equation}
    (\tilde{\mathbf Q})^{\lambda,\sigma}_{A,nlm}
    =\int_{\mathbb R^3} d\mathbf r\;
    (\chi^\lambda_A(\mathbf r))^{*}\,
    \chi^\sigma_A(\mathbf r)\,
    \tilde{\Phi}^{nlm}_A(\mathbf r),
\end{equation}

\noindent
where $\lambda,\sigma$ label primitive GTOs. These primitive-basis integrals depend only on the element $A$ (and the auxiliary indices) and can therefore be precomputed.

Given the contraction coefficients for a specific atom $A_i$, the corresponding contracted on-site three-index overlap integral is obtained by a straightforward expansion,

\begin{equation}
\begin{split}
    (\tilde{\mathbf Q})^{\mu,\nu}_{A_i,nlm}
    &=\int_{\mathbb R^3} d\mathbf r\;
    \Phi^\mu_{A_i}(\mathbf r)^{*}\,
    \Phi^\nu_{A_i}(\mathbf r)\,
    \tilde{\Phi}^{nlm}_A(\mathbf r) \\
    &=\sum_{\lambda,\sigma}
    \left(c^{\lambda}_{A_i}\right)^{*}
    c^{\sigma}_{A_i}
    \int_{\mathbb R^3} d\mathbf r\;
    (\chi^\lambda_A(\mathbf r))^{*}\,
    \chi^\sigma_A(\mathbf r)\,
    \tilde{\Phi}^{nlm}_A(\mathbf r) \\
    &=\sum_{\lambda,\sigma}
    \left(c^{\lambda}_{A_i}\right)^{*}
    c^{\sigma}_{A_i}\,
    (\tilde{\mathbf Q})^{\lambda,\sigma}_{A,nlm}.
\end{split}
\end{equation}

\noindent
Thus, while the contracted on-site three-index overlap integrals
$(\tilde{\mathbf Q})^{\mu,\nu}_{A_i,nlm}$ become atom-specific through the
environment-dependent contraction coefficients, they can be swiftly evaluated from the element-tabulated primitive integrals
$(\tilde{\mathbf Q})^{\lambda,\sigma}_{A,nlm}$ once the contraction coefficients are known (e.g., from the basis specification provided by the electronic-structure driver). The contraction coefficients are readjusted with respect to their Frobenius norm for consistent scales of features.

Reduced embeddings from each QMM are passed through linear layers with learnable weights, producing the initial hidden features $\textbf{h}^{t=0}_A$. The number of channels is listed in \SI{}Table \ref{tab:app:Nlp_each_lp}.

\paragraph{Message Passing}

At layer $t$, the equivariant message from atom $B$ to atom $A$, $\textbf{m}^t_{BA}$, is computed via block convolutions over the off-diagonal QMM blocks, following \cite{qiao_orbnet_equi_2022}. These blocks are projected onto convolution channels, with their number specified in \SI{}Table \ref{tab:app:hyperparameters}.

Messages from neighboring atoms are aggregated using multi-head attention. The resulting message for atom $A$ is:

\begin{equation}
    \tilde{\textbf{m}}^t_A=\sum_B\bigoplus_{i,j}\textbf{m}^{t,i}_{BA}\cdot\alpha^{t,j}_{AB},
\end{equation}

\noindent
where $i$ is the convolution channel index, and $\alpha^{t,j}_{AB}$ is the invariant attention of the atom $A$ to the atom $B$ for the $j$-th attention head.
The aggregated message $\tilde{\textbf{m}}^t_A$ is then coupled with the node representation of the atom $A$ at layer $t$, $\textbf{h}^t_A$ with the point-wise interaction module defined in \cite{qiao_orbnet_equi_2022}, which updates the node representation to $\textbf{h}^{t+1}_A$. 
Further details on the message constructions and the point-wise interaction can be found in \SI{}\appendixsection{} \ref{app:architecture}.

\paragraph{Atom-wise Decoding and Pooling}

The atom-wise decoding layer updates each node using its own features. A point-wise interaction is applied to $\textbf{h}^t_A$, producing the more abstract updated representation $\textbf{h}^{t+1}_A$.

After passing the $t_m$ message passing layers and $t_d$ atom-wise decoding layers, the final representation $\textbf{h}^{t_m+t_d}_A$ is used for the task-specific pooling operation. In this work, two distinct pooling operations are employed as in \cite{qiao_orbnet_equi_2022}: one for predicting the total energy and another for the FMO property.

The predicted total energy, $\hat{E}_\text{tot}$, is given by

\begin{equation}
\label{eqn:e_tot_pooling}
    \hat{E}_\text{tot}=\sum_A\textbf{W}_o\cdot\lVert\textbf{h}^{t_m+t_d}_A\rVert+b_{Z_A},
\end{equation}

\noindent
where $A$ is the atom index, $\textbf{W}_o$ is a learnable matrix, $b_{Z_A}$ is the element-wise energy bias, i.e., the element-wise shift, for atom $A$'s atomic number $Z_A$. The element-wise shifts are initialized from a linear regression of the total energy to atomic numbers. 

For predicting FMO energies, pooling with global attention is applied as follows:

\begin{equation}
    a_A=\frac{\textbf{W}_a\cdot\lVert\textbf{h}^{t_m+t_d}_A\rVert}{\sum_B\textbf{W}_a\cdot\lVert\textbf{h}^{t_m+t_d}_B\rVert},
\end{equation}

\begin{equation}
    \hat{E}_\text{FMO}=\sum_Aa_A(\textbf{W}_o\cdot\lVert\textbf{h}^{t_m+t_d}_A\rVert+b_{Z_A}),
\end{equation}

\noindent
where $a_A$ is a scalar attention to the atom $A$, and $\textbf{W}_a$ is a learnable matrix.

\subsection{Datasets}
\label{sec:datasets_and_trainings}

\paragraph{The OMol25 Dataset}

The OMol25 dataset is a large-scale collection comprising 140 million single-point calculations performed at the $\omega$B97M-V/def2-TZVPD level of theory \cite{levine2025openmolecules2025omol25}. It includes molecular systems spanning a wide range of charge states (0 to $\pm$10) and spin configurations (spin quantum numbers from 0 to 5, corresponding to up to 10 unpaired electrons and a maximum spin multiplicity of 11). Covering 83 elements and systems containing up to 350 atoms, the dataset captures substantial chemical diversity and complexity.

In this work, we use the 4M (4 million) subset of OMol25, which is uniformly sampled from the full 140M dataset. When employing g-xTB as the underlying semi-empirical quantum mechanical method, molecules containing lanthanides are excluded from both training and evaluation. This is due to the f-in-core strategy adopted in g-xTB \cite{froitzheim2025gxtb}, wherein f-electrons are treated as core electrons rather than valence electrons, leading to ambiguities in spin-state representation. Additionally, all molecules that failed to converge or produced errors during preprocessing are removed from the dataset.

The official test set of the OMol25 dataset is not publicly available with labels. Consequently, we designate the provided validation split, whose labels are accessible, as our in-house test set for model evaluation. To retain a validation set for hyperparameter tuning, we randomly sample 65,536 molecules from the original training split and use them for validation. After this re-partitioning, the dataset comprises 3.85M molecules for training, 65.5K for validation, and 2.70M for testing. The model was trained with the delta-learning strategy, without the element-wise or charge biases.

\paragraph{The \tlxsolv{} Dataset}

Transition1x is a dataset containing 10,073 reactions and their reaction pathways, including reactants, products, and transition states. Each reaction includes both converged and unconverged pathways from NEB calculations; therefore, the number of data points varies across reactions. In total, the dataset comprises 9.6M single-point calculations at the $\omega$B97x/6-31G(d) level of theory.

To build a reactive potential under different solvent conditions, we carefully selected geometries from the Transition1x dataset and used them for single-point calculations at the $\omega$B97M-V/def2-TZVPD level of theory. Specifically, we extracted eight geometries per reaction: (1) the relaxed reactant, (2) the relaxed product, (3) the converged transition state and its two neighboring images, and (4) three randomly selected images. This procedure yielded 80,584 geometries, which were then used for single-point calculations under four solvent conditions: vacuum, water, methanol, and toluene. Consequently, the dataset contains 322,336 data points with DFT-evaluated energies and forces. The single-point calculations were performed using ORCA 6.1.0 \cite{neese2025orca}, with solvation modeled using SMD \cite{marenich2009smd}. The `RIJ-COSX' and `TIGHTSCF' flags were enabled.

Given the current implementation of g-xTB 2.0.0 \cite{froitzheim2025gxtb}, we used the generalized Born model with finite dielectric constant (GBE) to generate the QMMs. The data points that SCF failed are removed from the data set for training the \ourmodel{} model. During fine-tuning of the \foundationmodel{} model, we additionally applied solvent-specific element-wise shifts.

\paragraph{Evaluation Datasets and Details}

QM9star is a quantum chemistry dataset comprised of about 2 million data points in total: $\sim$120K of neutral singlets (neutrals, $Q=0$, $S=0$), $\sim$435K of $+1$ charged singlets (cations, $Q=+1$, $S=0$), $\sim$721K of $-1$ charged singlets (anions, $Q=-1$, $S=0$), and $\sim$731K of neutral doublets (radicals, $Q=0$, $S=1/2$) \cite{tang2024qm9star}. Each datapoint is a small drug-like molecule optimized at the B3LYP-D3(BJ)/6-311+G(d,p) level of theory with subsequent calculations of quantum mechanical properties. 

In the QM9star dataset, anions, cations, and radicals are generated by removing a hydrogen atom from neutral “parent molecules.” At the site of hydrogen removal, subtracting an electron yields a singlet cation, adding an electron produces a singlet anion, and leaving the molecule with an unpaired electron results in a radical \cite{tang2024qm9star}.

To construct training subsets, we ensured that no parent molecule appears in more than one of the training, validation, or test sets. Validation sets are sized at 10\% of their corresponding training subsets, except for the full 400K training set, which uses a fixed 20K validation set. Additionally, each training and validation set of size $N$ is sampled to include equal numbers ($N/4$ each) of neutrals, radicals, cations, and anions.

While training all models (including \ourmodel{} and the baselines) on the QM9star dataset, we applied charge shifts to account for the overall effect of molecular charge. The charge shift, denoted by $b_Q$, is a learnable bias term added during the final pooling stage to represent the average contribution of each charge state.

Hence, the predicted energy of a molecule, $\hat{E}_\text{tot}$, from the models is given by

\begin{equation}
\label{eqn:charge_shift}
    \hat{E}_\text{tot}=\left(\sum_A\hat{E}_A+b_{Z_A}\right)+b_Q,
\end{equation}

\noindent

where, $A$ denotes the atom (node) index, $\hat{E}_A$ is the atomic energy predicted by the GNN, $b_{Z_A}$ is the element-wise energy bias (i.e., the shift associated with atom $A$'s atomic number $Z_A$), and $b_Q$ is the charge-dependent energy bias (i.e., the \textit{charge shift}). We initialize the charge shifts using the average total energy $E_\text{tot}$, minus the sum of element-wise biases for each charge state in the training set. Notably, applying the charge shift stabilized training and reduced both the frequency and magnitude of outliers for \ourmodel{}.

QMSpin is a dataset for the closed-shell and open-shell species prediction tasks, which consists of 4.9K singlet-optimized and 7.8K triplet-optimized carbene geometries. The singlet ($S=0$) and triplet ($S=1$) energies of each geometry is calculated, resulting in a total of 25.6K energy points. The geometries are optimized via restricted open-shell B3LYP/def2-TZVP, and the energies are obtained at MRCISD+Q-F12/cc-pVDZ-F12 \cite{schwilk2020qmspin}. 

The original geometries of polypeptides were obtained from the PEPCONF dataset \cite{prasad2019pepconf}, which consists of neutral ($Q=0$, $S=0$), anion ($Q=1$, $S=0$), and cation ($Q=-1$, $S=0$) polypeptide molecules. The polypeptide molecules are composed only of elements H, C, N, and O. Since \ourmodel{} predicts open-shell species properties, we created radical species by either removing an electron from an anion or adding an electron to a cation, to neutralize the charge.

After collecting the ground-state geometry of each molecule from the PEPCONF dataset, which then underwent three consecutive geometry optimizations: first with GFN2-xTB \cite{bannwarth2019gfn2}, followed by B3LYP-D3(BJ)/def2-SVP and then B3LYP-D3(BJ)/6-311+G(d,p).

The single point total energies $E_\text{tot}$ of polypeptides are the prediction targets, specifically with 58 neutrals, 46 radicals, 25 cations, and 22 anions. The number of heavy atoms of the QM9star dataset ranges from 1 to 9, whereas it ranges from 16 to 36 in the polypeptide dataset, making it a suitable dataset for evaluating the size extrapolation. 

Hessian QM9 is an implicit solvation dataset comprising 41.6K molecules in four solvent environments--vacuum, water, tetrahydrofuran, and toluene--distinguished by their dielectric constants ($\epsilon_r$). Each datapoint is computed at the $\omega$B97x/6-31G* level of theory using the solvation model based on density (SMD) \cite{marenich2009smd}. For generating QMMs, we use the CPCM method \cite{barone1998cpcm1} implemented in tblite \cite{tblite} with the same dielectric constants used for Hessian QM9.

\subsection{Experiments}
\label{sec:method_experiments}

\paragraph{Nudged Elastic Band (NEB)}

All NEB simulations are performed using the Atomic Simulation Environment (ASE) \cite{hjorth2017ase}. The FIRE optimizer is used for NEB optimization \cite{bitzek2006fire}. Each simulation uses nine images in total, including the reactant, product, and seven intermediate images. The initial path is first relaxed with NEB until the maximum force ($f_\text{max}$) is below 0.20 eV/\AA, followed by climbing-image NEB (CI-NEB) optimization with an $f_\text{max}$ threshold of 0.05 eV/\AA. The reactant and product geometries are optimized primarily using the Broyden--Fletcher--Goldfarb--Shanno (BFGS) method, and structures that do not converge are optimized using the FIRE optimizer. All geometry optimizations are performed with the ASE implementation using a convergence threshold of $f_\text{max}=0.02$ eV/\AA. The relative energies of the reactant, product, and transition state are then used to calculate the reaction barrier, $\Delta E^\ddagger=E_\text{TS}-E_\text{R}$, and reaction energy, $\Delta_r E=E_\text{P}-E_\text{R}$, where $E_\text{TS}$, $E_\text{R}$, and $E_\text{P}$ denote the transition-state, reactant, and product energies, respectively.

\paragraph{Umbrella Sampling}

We performed explicit-solvent umbrella sampling with UMA-s-1p2 to establish a strong baseline for explicit-solvation reaction modeling. UMA was selected as a competitive MLIP, providing a favorable reference point for explicit-solvent simulations. Direct TS discovery from unbiased explicit-solvent molecular dynamics (MD) is impractical for the Claisen rearrangement because barrier-crossing events are expected to be rare on accessible MD timescales, considering the high barrier height. We therefore used a TS-informed, or oracle, umbrella-sampling in which transition-state-like windows were initialized from reference structures calculated at $\omega$B79M-V/def2-TZVPD/SMD. This setup gives UMA substantial prior information and should be viewed as a favorable lower-bound estimate of the computational burden required for explicit-solvent reaction profiling.

The initial explicit-solvent configurations for the reactant-, TS-, and product-like umbrella windows are generated using PACKMOL~\cite{martinez2009packmol} by embedding the corresponding solute geometries in cubic solvent boxes of size $25~\text{\AA} \times 25~\text{\AA} \times 25~\text{\AA}$, reducing solute--image interactions under periodic boundary conditions. Umbrella sampling is performed with UMA-s-1p2 using the collective variable $\xi = r_{\mathrm{CC}} - r_{\mathrm{CO}}$, where $r_{\mathrm{CC}}$ and $r_{\mathrm{CO}}$ denote the forming C--C and breaking C--O bond distances, respectively. We sample 31 umbrella windows with centers ranging from $\xi = -1.5~\text{\AA}$ to $\xi = +1.5~\text{\AA}$ at intervals of $0.1~\text{\AA}$. A harmonic restraint force constant of $10.0~\mathrm{eV}/\text{\AA}^{2}$ is used for all windows. Each window is simulated in the NVT ensemble at $300~\mathrm{K}$ for $15~\mathrm{ps}$, consisting of $5~\mathrm{ps}$ equilibration and $10~\mathrm{ps}$ production, with a timestep of $0.5~\mathrm{fs}$. For windows centered between $\xi = -0.6~\text{\AA}$ and $\xi = +0.6~\text{\AA}$, the solvated TS structure is used as the initial configuration, whereas reactant- and product-solvated structures are used for windows with $\xi > +0.6~\text{\AA}$ and $\xi < -0.6~\text{\AA}$, respectively. The biased distributions from the production trajectories are reweighted and combined using the weighted histogram analysis method (WHAM) to reconstruct the one-dimensional potential of mean force along $\xi$, from which the activation barrier and reaction free energy are estimated.

\paragraph{Wall time measurement}

The wall times of the reaction pathway predictions for each method are measured using a single NVIDIA H200 GPU and 32-cores of AMD EPYC 9554 @ 3.1 GHz. 

\backmatter

\bmhead{Data Availability}

All datasets used in this study will be made available upon publication.

\bmhead{Code Availability}

All code used in this study will be made available upon publication.

\bmhead{Acknowledgements}

B.S.K. acknowledges graduate research funding from the California Institute of Technology, support from the Pritzker AI+Science fund and the Eddleman Graduate Fellowship. W.A.G. acknowledges support from NSF(CBET-231117). A.A. acknowledges support from the Bren endowed chair, ONR (MURI grant N00014-23-1-2654), and the Schmidt Sciences AI2050 senior fellow program.  This work used the Delta system at the National Center for Supercomputing Applications through allocation DMR160114 from the Advanced Cyberinfrastructure Coordination Ecosystem: Services \& Support (ACCESS) program, which is supported by National Science Foundation grants \#2138259, \#2138286, \#2138307, \#2137603, and \#2138296. B.S.K. acknowledges Robert Kalescky, John Santerre, and Bivin Sadler for their help in organizing the computational resources used in this research through SMU’s O’Donnell Data Science and Research Computing Institute.

\bibliography{bibliography.bib}

\ifthenelse{\boolean{arxiv}}
{
\newpage
\appendix
\renewcommand{\thesection}{S\arabic{section}}
\renewcommand{\theequation}{S\arabic{equation}}
\renewcommand{\thefigure}{S\arabic{figure}}
\renewcommand{\thetable}{S\arabic{table}}

\section{Architecture}
\label{app:architecture}

\paragraph{Equivariant Normalization}

The equivariant normalization ($\text{EvNorm}$) used in the UNiTE framework is defined by:

\begin{gather}
    (\overline{\textbf{h}},\hat{\textbf{h}})=\text{EvNorm(\textbf{h})},\\
\overline{\textbf{h}}_{nlp}:=\frac{\lVert\textbf{h}_{nlp}\rVert-\mu^h_{nlp}}{\sigma^h_{nlp}},\\
    \lVert\textbf{h}_{nlp}\rVert:=\sqrt{\sum_m h^2_{nlpm}+\epsilon^2}-\epsilon,\\
    \hat{h}_{nlpm}:=\frac{h_{nlpm}}{\lVert\textbf{h}_{nlp}\rVert+1/\beta_{nlp}+\epsilon}.
\end{gather}

\noindent
Here, $\lVert\textbf{h}_{nlp}\rVert$ is the invariant content of $\textbf{h}$, $\mu^h_{nlp}$  and $\sigma^h_{nlp}$ are the mean and variance of $\lVert\textbf{h}\rVert$, respectively, $\epsilon$ is a stability factor and $\beta_{nlp}$ is a positive learnable scalar that controls the amount of $\lVert\textbf{h}_{nlp}\rVert$ information in the vector part.

\paragraph{Creating Messages by Block-wise Convolutions}

The block-wise convolutions characterize the message passing procedure of the UNiTE framework \cite{qiao_orbnet_equi_2022}. Supposing we are using one feature matrix $\textbf{O}$, a message from the atom $A$ to the atom $B$ for the $i$-th convolution channel is defined by:

\begin{equation}
\label{eqn:message}
    \textbf{m}_{AB,\nu}^i=\sum_\mu(\rho_i(\textbf{h}_A))_\mu (\textbf{O})^{\mu,\nu}_{AB},
\end{equation}

\noindent
where $\mu=(n_1,l_1,m_1)$ and $\nu=(n_2,l_2,m_2)$ are atomic orbital indices, $\textbf{h}_A$ is the hidden representation of the atom $A$, and $\rho$ is a matching layer defined by:

\begin{equation}
    (\rho_i(\textbf{h}_A))_\mu=\text{Gather}(\textbf{W}_l^i\cdot(\textbf{h}_A)_{l(p=+1)m},n[\mu,Z_A]).
\end{equation}

\noindent
Here, $\textbf{W}^i_l$ is a learnable weight matrix for the $i$-th convolution channel specific to angular momentum $l$. The ``$\text{Gather}$'' operation rearranges and maps the input hidden features to the valid orbital index $\mu$ using the principal quantum number $n$ of atomic orbital $\mu$ for the atom $Z_A$.

The messages are aggregated with multi-head attention as in Equation (\textcolor{red}{17}), creating an aggregated message $\tilde{\textbf{m}}_{A}$ (for the atom $A$) that is to be used for updating the hidden representation of the atom $A$. The aggregated message is passed to a reverse matching layer $\rho^\dagger$, defined by:

\begin{equation}
    (\rho^\dagger(\tilde{\textbf{m}}_A))_{lpm}=\begin{cases}
        \textbf{W}^\dagger_l\cdot\sum_\mu\text{Scatter}((\tilde{\textbf{m}}_{A})_\mu,n[\mu,Z_A]),\qquad &p=+1,\\
        0, &p=-1,
    \end{cases}
\end{equation}

\noindent
where $\textbf{W}^\dagger_l$ is a learnable weight matrix specific to $l$, and the ``$\text{Scatter}$'' operation flattens the message $\tilde{\textbf{m}}_{A}$ using $n[\mu,Z_A]$. The combined operation maps the message and projects into the same dimension as the atomic hidden representation $\textbf{h}_A$.

\paragraph{The Point-wise Interaction}

The point-wise interaction module, $\phi$, performs normalization and applies non-linearity. It couples two equivariant features, $\textbf{h}$ and $\textbf{g}$, to produce an updated representation $\textbf{h}' = \phi(\textbf{h}, \textbf{g})$ through the following operations:

\begin{gather}
    (\overline{\textbf{h}},\hat{\textbf{h}})=\text{EvNorm(\textbf{h})},\\
    \textbf{f}_{lpm}=(\text{MLP}_1(\overline{\textbf{h}}))_{lp}\odot(\hat{\textbf{h}}_{lpm}\cdot\textbf{W}^{\text{in}}_{l,p}),\\
    \label{eqn:app:cg_coupling_decoding}\textbf{q}_{lpm}=\textbf{g}_{lpm}+\sum_{l_1,l_2}\sum_{m_1,m_2}\sum_{p_1,p_2}\textbf{f}_{l_1p_1m_1}\cdot\textbf{g}_{l_2p_2m_2}\cdot C^{lm}_{l_1m_1,l_2m_2}\cdot\delta^{(-1)^{l_1+l_2+l}}_{p_1\cdot p_2\cdot p},\\
    (\overline{\textbf{q}},\hat{\textbf{q}})=\text{EvNorm(\textbf{q})},\\
    \textbf{h}'_{lpm}=\textbf{h}_{lpm}+(\text{MLP}_2(\overline{\textbf{q}}))_{lp}\odot(\hat{\textbf{q}}_{lpm}\cdot\textbf{W}^{\text{out}}_{l,p}),
\end{gather}

\noindent
where $\odot$ symbol indicates an element-wise product, $\text{MLP}_1$ and $\text{MLP}_2$ are multi-layer perceptron (MLP) layers with depth and activation functions specified in Table \ref{tab:app:hyperparameters}, $\textbf{W}^{\text{in}}_{l,p}$ and $\textbf{W}^{\text{out}}_{l,p}$ are learnable weight matrices acting on each $(l,p)$, $C^{lm}_{l_1m_1,l_2m_2}$ is the Clebsch-Gordan coefficient, and $\delta^i_j$ is the Kronecker delta function.

\paragraph{Updating Hidden Representations}

A message passing layer updates the hidden representation $\textbf{h}^t_A$ of the atom $A$ with the reverse-matched aggregated message $\rho^\dagger(\tilde{\textbf{m}}^t_A)$ by:

\begin{equation}
    \textbf{h}^{t+1}_A=\phi(\textbf{h}_A^t,\rho^\dagger(\tilde{\textbf{m}}^t_A)),
\end{equation}

\noindent
where $\textbf{h}^{t+1}_A$ is the updated hidden representation.

An atom-wise decoding layer decodes atom-wise information with local interactions, which is,

\begin{equation}
    \textbf{h}^{t+1}_A=\phi(\textbf{h}^t_A,\textbf{h}^t_A).
\end{equation}

When training \ourmodel{} on the OMol25 dataset, we found that the quadratic scaling induced by the self Clebsch-Gordan coupling in Eq.~\ref{eqn:app:cg_coupling_decoding} led to instability in training when multiple decoding layers are stacked, particularly given the diversity of the dataset. To mitigate this issue, we skip the Clebsch-Gordan coupling step during training on OMol25. In equation, we set $\mathbf{q}_{lpm} = \mathbf{f}_{lpm}$.

\section{Training}
\label{app:training}

\subsection{Workflow}

During inference time, the model requires running semi-empirical QM calculations using spGFN1-xTB \cite{neugebauer_spgfnxtb_2023} or g-xTB \cite{froitzheim2025gxtb} for each molecule. However, during training, since the dataset consists of a fixed set of molecules that are used repeatedly, we preprocess all molecules by performing the semi-empirical QM calculations in advance. The resulting features are then used directly during the GNN training phase.

\begin{algorithm}
\caption{Training Procedure}\label{alg:training_procedure}
\begin{algorithmic}
\item\texttt{\# 1. Creating orbital features\\}
\For{$x_i$, $y_i$\,\texttt{in}\,$\mathcal{D}_\text{train}$} \Comment{$x_i$: molecular information, $y_i$: label, $\mathcal{D}_\text{train}$: train dataset}
\State $\textbf{T}_i\gets$\texttt{xTB}($x_i$) \Comment{$\textbf{T}_i$: orbital features}
\State Add $\textbf{T}_i$ to $\mathcal{D}_\text{train}$
\EndFor
\For{$x_j$, $y_j$\,\texttt{in}\,$\mathcal{D}_\text{test}$} \Comment{$\mathcal{D}_\text{test}$: test dataset}
\State $\textbf{T}_j\gets$\texttt{xTB}($x_j$)
\State Add $\textbf{T}_j$ to $\mathcal{D}_\text{test}$
\EndFor
\item\texttt{\\ \# 2. Train and test neural network\\}
\item$\mathcal{F}\gets$\texttt{fit}($\mathcal{F},\mathcal{D}_\text{train}$) \Comment{$\mathcal{F}$: UNiTE}
\item$\hat{\textbf{y}}_\text{test}\gets$\texttt{predict}($\mathcal{F},\mathcal{D}_\text{test}$) \Comment{$\hat{\textbf{y}}_\text{test}$: test set predictions}
\end{algorithmic}
\end{algorithm}

\subsection{Training Configuration}

All orbital features are generated either using the tblite package \cite{tblite} with spGFN1-xTB \cite{neugebauer_spgfnxtb_2023} or the binary executable of g-xTB \cite{froitzheim2025gxtb}. For the experiments in Sections \ref{sec:qm9star_spin_charge} and \ref{sec:env_effects}, we used the default hyperparameters used in the previous OrbNet-Equi work \cite{qiao_orbnet_equi_2022} (`Small' in Table \ref{tab:app:hyperparameters}). In contrast, for the large-scale experiments in Section \ref{sec:foundation_model}, several hyperparameters are modified to increase model capacity (`Large' in Table \ref{tab:app:hyperparameters}).

We used the Adam optimizer \cite{kingma_adam_2014} with a learning rate schedule consisting of a linear warm-up phase followed by cosine annealing. The corresponding hyperparameters are detailed in Table \ref{tab:app:hyperparameters}. Additionally, the smoothL1Loss loss function was used \cite{SmoothL1Loss}, which is a loss function with a quadratic slope below a certain threshold and a linear slope above the threshold.

\begin{table}[ht]
\centering
\caption{Hyperparameters for \ourmodel{} used for experiments.}
\label{tab:app:hyperparameters}
\begin{tabular}{lcc}
\hline
Description                                   & Small & Large         \\ \hline
Node hidden dimension &    256   &    256       \\
Number of channels for each ($l,p$)                & Shown in Table \ref{tab:app:Nlp_each_lp}         & Shown in Table \ref{tab:app:Nlp_each_lp} \\
Number of message-passing update steps        & 4   & 8           \\
Number of point-wise decoding steps           & 4    & 4             \\
Number of convolution channels                & 8     & 16             \\
Number of attention heads                     & 8   & 16             \\
Depth of MLPs                                 & 2    & 2             \\
Activation function                           & Swish   & Swish         \\
Number of radial basis functions              & 16     & 32            \\
Stability factor $\epsilon$ in EvNorm layers     & 0.1      & 0.1          \\
Max neighbors     & 64      & 64          \\
Maximum learning rate     & 0.0005     & 0.0008      \\
Warm-up ratio   & 0.333 &  0.0835         \\
Number of epochs     & 300    & 80       \\
Batch size     & 64    & 128       \\
Stable decoding layer & False & True \\
Max gradient norm & - & 100.0 \\
Feature normalization & - & LayerNorm \\
Encoding normalization & BatchNorm & LayerNorm \\
Message-passing node normalization & - & LayerNorm \\
Message-passing normalization & LayerNorm & LayerNorm \\
Decoding normalization & BatchNorm & LayerNorm \\
Max gradient norm & - & 100.0 \\
Total number of learnable parameters                    & 2.1M    & 7.5M     \\ \hline
\end{tabular}
\end{table}

\begin{table}[ht]
\centering
\caption{Number of channels for each ($l,p$).}
\label{tab:app:Nlp_each_lp}
\begin{tabular}{cccccc}
\hline
 &  $l=0$   &  $l=1$  &  $l=2$  &  $l=3$  &  $l=4$ \\ \hline
$p=+1$ & 128 & 48 & 24 & 12 & 6 \\
$p=-1$ & 24  & 8  & 4  & 2  & 0 \\ \hline
\end{tabular}
\end{table}

\subsection{The OMol25 Dataset}

For training \ourmodel{} on the OMol25 dataset, we employ the g-xTB framework \cite{froitzheim2025gxtb}. As the current implementation of g-xTB is distributed as a binary executable, direct access to orbital-level quantities, such as the Fock, density, overlap, and core Hamiltonian matrices, is not readily available. To address this limitation, we reconstruct the orbital features from the wavefunction exported in Molden format, which provides access to the converged electronic structures. Using the parsed molecular orbital coefficients, we then build the required orbital features with the PySCF package \cite{sun2020pyscf}.

During the construction of the core Hamiltonian matrices using the PySCF package, we observed that their magnitudes were significantly larger than those obtained from spGFN1-xTB as implemented in the tblite package. This difference resulted in feature matrices with widely varying numerical scales, which introduced instability during training. Therefore, to ensure consistency in feature magnitudes across different orbital features, we normalize the core Hamiltonian by the total nuclear charge of the system. Specifically, for molecule $i$, we define

\begin{equation}
\tilde{\mathbf{H}}_{\text{core},i}
=
\frac{\mathbf{H}_{\text{core},i}}{\sum_{A_i} Z_{A_i}},
\end{equation}

\noindent
where $\sum_{A_i} Z_{A_i}$ is the total number of protons in the molecule, and $\tilde{\mathbf{H}}_{\text{core},i}$ denotes the normalized core Hamiltonian matrix. The normalized core Hamiltonian matrix is used with other orbital features during training and evaluating \ourmodel{} on the OMol25 dataset.

\subsection{The \tlxsolv{} Dataset}

For fine-tuning the pretrained \ourmodel{} model on the \tlxsolv{} dataset, we unfreeze all model parameters and train for 150 epochs using the same learning rate as in pretraining, 0.0008. We use the GBE (generalized Born with finite epsilon) implementation in g-xTB v2.0.0 \cite{froitzheim2025gxtb}.

During inference, because only relative energies are relevant, we use the vacuum element-wise shifts for unseen solvents. As a result, absolute energy predictions for unseen solvents are expected to be inaccurate.

\subsection{QM9star}
\label{app:qm9star_training_detail}

The QM9star dataset consists of neutral singlets (neutrals), neutral doublets (radicals), $+1$ charged singlets (cations), and $-1$ charged singlets (anions). The molecular properties are calculated at the B3LYP-D3(BJ)/6-311+G(d,p) level of theory \cite{stephens1994b3lyp, becke1988becke88, lee1988lyp}, where 36 or 39 properties are available for each molecule depending on its spin state, including internal energy ($U_0$), heat capacity ($C_v$), HOMO energy, LUMO energy, zero-point vibrational energy (ZPVE), dipole moment ($\mu$), and isotropic polarizability ($\alpha$). Although the dataset also contains force labels, the magnitudes are practically too tiny for any evaluation to be performed. Hence, we skip the force predictions for the open-shell or charged molecules task for this work.

\begin{figure}[ht]
    \centering
    \includegraphics[width=1\linewidth]{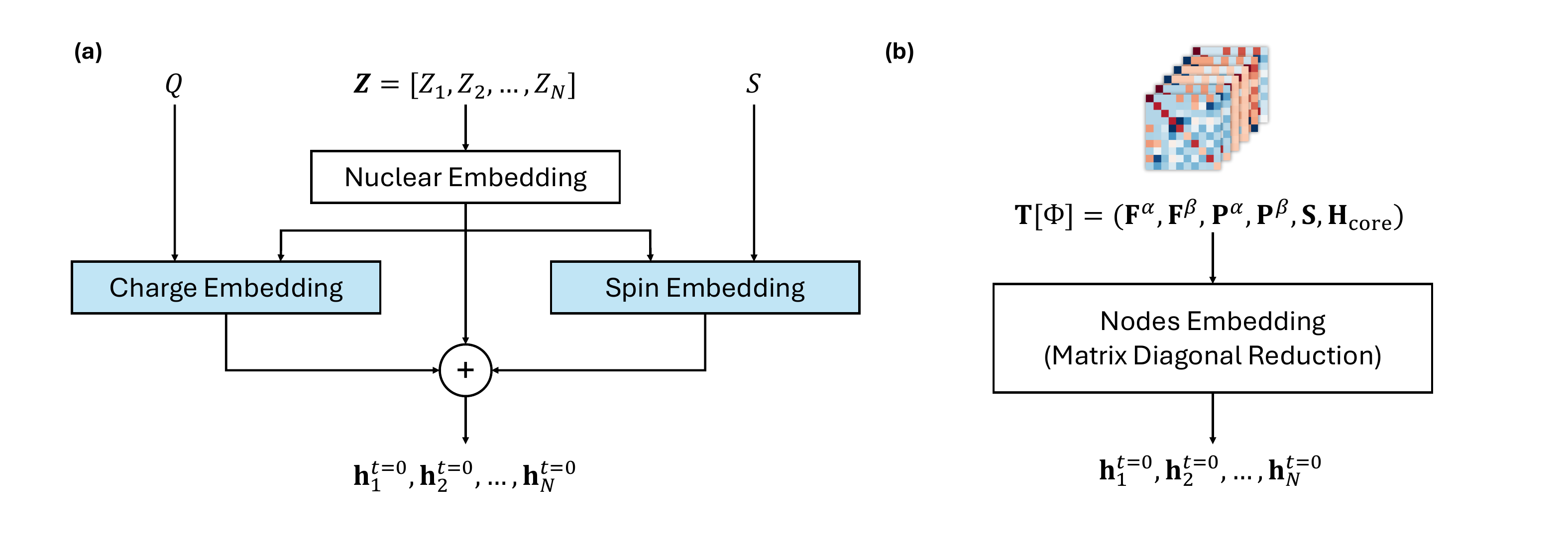}
    \caption{(a) SpookyNet embedding style for embedding spin and charge information along with the atomic numbers. We apply the SpookyNet embeddings to all the models except for \ourmodel{} to incorporate spin and charge information \cite{unke_spookynet_2021}. (b) \ourmodel{} embedding style for embedding electronic structure information, which contains the spin and charge information inherently.}
    \label{fig:app:different_embedding_styles}
\end{figure}

Except for SpookyNet \cite{unke_spookynet_2021} and EquiformerV2 \cite{liao2023equiformerv2}, all models were implemented and trained using the Geom3D benchmark repository \cite{liu2023geom3d}, including SchNet \cite{schutt2018schnet}, PaiNN \cite{schutt2021painn}, DimeNet++ \cite{gasteiger2022dimenetpp}, and NequIP \cite{batzner2022nequip}. For EquiformerV2, we used the same hyperparameters as in their QM9 experiments \cite{ramakri_qm9_2014, liao2023equiformerv2}.

In all models, element-wise biases (or shifts) were initialized via linear regression using only the corresponding training set in each run. The QM9star dataset consists of molecules with total charges of +1, 0, and -1. Since our strategy for initializing the charge shift does not allow other charges than those in the training set distribution, the extrapolation to out-of-distribution charged molecules (e.g., +2, -2, etc.) is expected to fail. This issue can be fixed either by preparing species with various charges in the training set or by implementing appropriate modeling of the charge biases, i.e., the average charge effects.

During training, validation errors were monitored, and the model checkpoint with the lowest validation error was selected for testing. Validation frequencies followed those specified in the original papers (e.g., SpookyNet was validated every 1,000 mini-batches \cite{unke_spookynet_2021}). Early stopping with a patience of 150 epochs was applied to all models unless otherwise specified. Models that failed to converge were excluded from the reported results.

For training the models other than \ourmodel{}, we use the SpookyNet embedding style \cite{unke_spookynet_2021} as their embeddings so that the different models can accept spin and charge information. The difference in the embeddings of the graph nodes is shown in Figure \ref{fig:app:different_embedding_styles}. The models were trained on one of the A100 or GH200 GPUs.

\subsubsection{Ablation Studies}

With the QM9star dataset, we conducted ablation studies on \ourmodel{}. We used the 40K subset training set to assess the effects of the different factors. The results of the ablation studies are presented in Table \ref{tab:app:ablation_studies}. Details on different methods are described below.

\begin{table}[ht]
\centering
\tiny
\caption{Ablation studies on \ourmodel{}. The numbers are mean absolute errors (MAEs) on the 40K QM9star training subset.}
\label{tab:app:ablation_studies}
\begin{tabular}{cccccc}
\hline
Methods                                          & All   & Neutral & Radical & Cation & Anion \\ \hline
Default                                          & 21.69 & 12.52   & 16.80   & 27.90  & 27.37 \\
Layer-wise Aggregation                            & 22.16 & 12.97   & 17.45   & 28.08  & 27.99 \\
Empirical Physical Interactions & 27.60   & 12.57   &  17.97  & 37.86   & 38.42 \\
Layer-wise Aggregation + Empirical Physical Interactions & 28.50 & 13.01   & 18.97   & 38.58  & 39.77 \\
Attention Renormalization (Node Norm)  & 20.87 & 11.79   & 16.22   & 26.72  & 26.61 \\
Attention Renormalization (Layer Norm)  & 20.97 & 12.23   & 16.51  & 26.80  & 26.27 \\
\hline
\end{tabular}
\end{table}

\paragraph{Layer-wise Aggregation}

Layer-wise aggregation in this study refers to the architectural consideration of whether to aggregate contributions from each layer or to use the result from the final layer. If layer-wise aggregation is applied, information from each layer is aggregated, which is pooled by a task-specific operation.

First, we generalize the message passing and decoding procedures. After each $t$-th message passing layer, $T_\text{dec}^t$ subsequent decoding layers are applied. With an expression of $T^t_\text{dec}$ as a length $t$ vector, $T_\text{dec}=[T^1_\text{dec},T^2_\text{dec},...,T^t_\text{dec}]$, the OrbNet-Equi architecture with four message passing layers followed by four decoding layers can be expressed as $T_\text{dec}=[0,0,0,4]$.

In this ablation study, we consider $T_\text{dec}=[1,1,1,1]$, i.e., one decoding layer per message passing layer. This setting makes the total number of decoding layers the same as the original OrbNet-Equi architecture \cite{qiao_orbnet_equi_2022}. 

\paragraph{Empirical Physical Terms}

Inspired by SpookyNet \cite{unke_spookynet_2021} and FENNIX \cite{ple2023fennix}, we consider adding empirical physical terms, especially the electrostatic correction, $E_\text{elec}$, and the dispersion correction, $E_\text{disp}$. We adopted SpookyNet's implementation in \ourmodel{}, where the dispersion correction is the D4 dispersion correction of Grimme et al. \cite{caldeweyher2019d4dispersion}. The calculations of both terms require the prediction of atomic partial charges, which is performed by a neural network.

Atomic partial charges can also be obtained from spGFN1-xTB. Hence, delta-learning can be employed, which means that the neural network predicts $\Delta q_A=q_A+q_{A,\text{spGFN1-xTB}}$ for atom $A$. The atomic partial charges shall add up to the total charge of molecules, i.e.,

\begin{equation}
    Q=\sum_A q_A,
\end{equation}

\noindent
which applies the same to the atomic partial charges from spGFN1-xTB. Therefore, the sum of delta-partial charges adds up to zero, $\sum_A\Delta q_A=0$. To enforce this as a hard physical constraint, the predicted delta partial charges, $\Delta\hat{q}_A$, are obtained by following,

\begin{equation}
    \Delta\hat{q}_A=\hat{q}_A-\frac{1}{N_\text{atom}}\left(\sum_B\hat{q}_B\right),
\end{equation}

\noindent
where $\hat{q}_A$ is the predicted partial charge of the atom $A$ from the neural network, and $N_\text{atom}$ is the number of atoms in the molecule. The atomic partial charges are then used for calculations of the electrostatic and dispersion correction terms. The details of implementation can be found in  \cite{unke_spookynet_2021}.

\paragraph{Attention Renormalization}

We apply attention renormalization proposed by Liao et al. \cite{liao2023equiformerv2}. The ablation studies of EquiformerV2 suggested that attention re-normalization can bring some improvements in the performance. Similar to EquiformerV2, an MLP-based attention is used in \ourmodel{} and OrbNet-Equi. Each attention head at layer $t$, $\alpha^t_{AB}$, for node $A$ attending to the message from node $B$ is created from a two-layer MLP. When attention renormalization is applied, the features for generating the attention are normalized with a layer norm. Hence, the normalized attention is

\begin{equation}
    \alpha^t_{AB}=\sigma(\text{Linear}(\sigma(\text{Linear}(\text{Norm}(x))))),
\end{equation}

\noindent
where $\sigma(\cdot)$ is an activation function, $\text{Linear}(\cdot)$ is a linear function with learnable weights and biases, $\text{Norm}(\cdot)$ is a normalization layer, and $x$ is primitive information for attention.

\subsection{QMSpin}

SpookyNet \cite{unke_spookynet_2021}, MOB-ML \cite{cheng_MOBML_2022}, and TensorNet \cite{simeon2025tensornet_spincharge} evaluated their models using this dataset with a 20K training set and a 1K validation set.

For calculating the adiabatic spin gaps, molecules with both singlet-optimized and triplet-optimized geometries are required. As in Cheng et al., \cite{cheng_MOBML_2022}, we randomly sampled 1K molecules with singlet- and triplet-optimized geometries (2K geometries in total). Each geometry has singlet and triplet energies, so the test set effectively has 4K data points. Similarly, 250 molecules with singlet- and triplet-optimized geometries (500 geometries, 1K data points in total) were randomly selected to create the validation set. Finally, 20K data points were randomly sampled from all remaining data points to generate the training set.

\subsection{Hessian QM9}

The Hessian QM9 dataset is a dataset that contains 41.6K molecules in four different solvents: vacuum, water, toluene, and tetrahydrofuran (THF). The molecules are all closed-shell, neutral species. Therefore, this prediction task is possible with the orbital features set $\textbf{T}=(\textbf{F}, \textbf{P}, \textbf{S}, \textbf{H}_\text{core})$. However, for consistency throughout this work, we use the orbital features set $\textbf{T}=(\textbf{F}^\alpha, \textbf{F}^\beta, \textbf{P}^\alpha, \textbf{P}^\beta, \textbf{S}, \textbf{H}_\text{core})$. Note that these are closed-shell species, and therefore $\textbf{F}^\alpha=\textbf{F}^\beta$ and $\textbf{P}^\alpha=\textbf{P}^\beta$.

For feature generation, we use the ALPB method \cite{sigalov2006alpb, ehlert2021alpb2} implemented in the tblite package \cite{tblite}, with the same dielectric constants for labels \cite{williams2025hessianqm9}. We decided to use the ALPB method based on the result of the ablation study between different solvation methods as shown in Table \ref{tab:app:hessian_qm9_diff_solvents}. For vacuum, no implicit solvation model was used in the calculation. The usage of the implicit solvation model perturbs the electronic structure and the orbital features of molecules. As a visualization, Figure \ref{fig:app:hessian_qm9_solvent_comparison} displays an example of the perturbation from using the different solvents, where it shows the differences between the density matrices at different solvents to those at vacuum.

We split the dataset into a 32K molecules training set, a 3.2K molecules validation set for training, and the rest ($\sim$6.4K molecules) for testing. Since each molecule has energies for the four solvents, the effective total number of data points is 128K for training, 12.8K for validation, and $\sim$25.8K for testing. 

\begin{figure}[ht]
    \centering
    \includegraphics[width=1.0\linewidth]{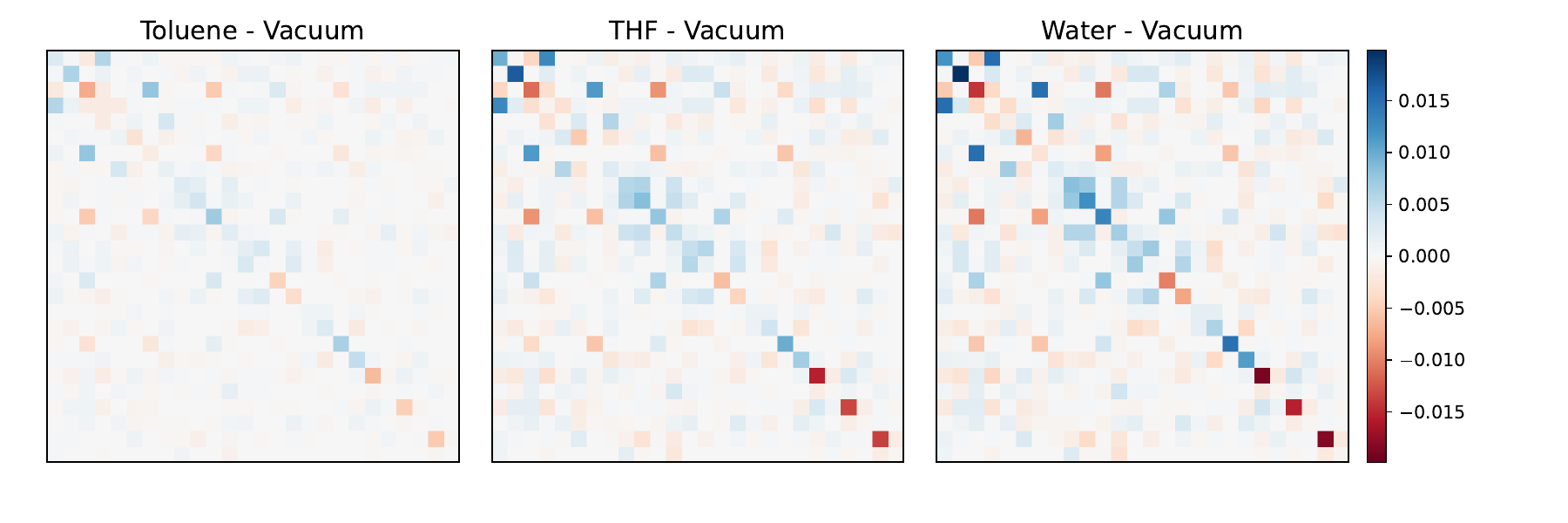}
    \caption{Density matrices difference for a sample molecule in different solvents. For example, the ``Toluene - Vacuum'' figure illustrates the matrix $\mathbf{P}_\text{Toluene}-\mathbf{P}_\text{Vacuum}$. The density matrices are generated using GFN1-xTB with ALPB solvation.}
    \label{fig:app:hessian_qm9_solvent_comparison}
\end{figure}

\subsection{The Polypeptide Dataset}

We sampled the initial atomic coordinates of the polypeptides from the PEPCONF dataset \cite{prasad2019pepconf}. The original PEPCONF dataset consists of $\sim$3.8K relative conformational energy data points calculated at the LC-$\omega$PBE-XDM/auc-cc-pVTZ level of theory. Since the QM9star dataset labels are calculated at the B3LYP-D3(BJ)/6-311+G(d,p) level of theory, we reoptimized the geometries and calculated the single point energy ($E_\text{tot}$) at the same level of theory.

Geometry optimizations of polypeptides were performed using the Orca 6.0 software \cite{neese2025orca}. The single point calculations for the optimized geometries were performed at the B3LYP-D3(BJ)/6-311+G(d,p) \cite{becke1988becke88, lee1988lyp, stephens1994b3lyp, grimme2010dftd} level of theory using the Psi4 software \cite{smith2020psi4}, with the default convergence thresholds, given by $10^{-6}$ for both energy and density, and $10^{-12}$ for integrals.

\subsection{QM9star in Random Uniform External Electric Fields}

From the original QM9star dataset \cite{tang2024qm9star}, we collect 10K training, 1K validation, and 2K test data points to study the behavior of \ourmodel{} in random uniform external electric fields. With the data points, we created 3 distinct datasets. First, the dataset with no electric field applied. The task is hence to predict the unperturbed single-point energies. For this dataset, we collected the single-point energies provided by the original QM9star dataset. Second, the dataset has uniform external electric fields in random directions and magnitudes, with a maximum magnitude of 0.005 atomic units (au).  Finally, the dataset has uniform external electric fields in random directions and magnitudes, with a maximum magnitude of 0.0005 au. To obtain the single point energies with uniform external electric fields, we used the psi4 software \cite{smith2020psi4}, and created the labels at the B3LYP-D3(BJ)/6-311+G(d,p) level of theory, which is the same level as the QM9star dataset. Note that we excluded radicals ($Q=0$, $S=1/2$) and collected only neutrals ($Q=0$, $S=0$), anions ($Q=-1$, $S=0$) and cations ($Q=+1$, $S=0$).

\subsection{Inference Time}

Inference times for all machine learning models were measured on an NVIDIA RTX 4090 GPU, using the mean single-batch inference time as the reported metric. For the QM9star calculation times, we sampled 500 randomly selected molecules using Psi4 \cite{smith2020psi4}. For measuring spGFN1-xTB using tblite \cite{tblite} and the QM9star calculation times, an AMD Ryzen 5600H CPU was used. For polypeptide dataset molecules, computations were performed on two 16-core Intel Skylake 2.1GHz CPUs. Notably, for the 500 QM9star molecules, the execution times on the Ryzen 5600H were found to be comparable to those on the Skylake CPUs.

\section{Additional Results}

\subsection{The OMol25 Dataset}
\label{sec:app:omol25_result_additional}

\begin{figure}[!ht]
    \centering
    \includegraphics[width=1\linewidth]{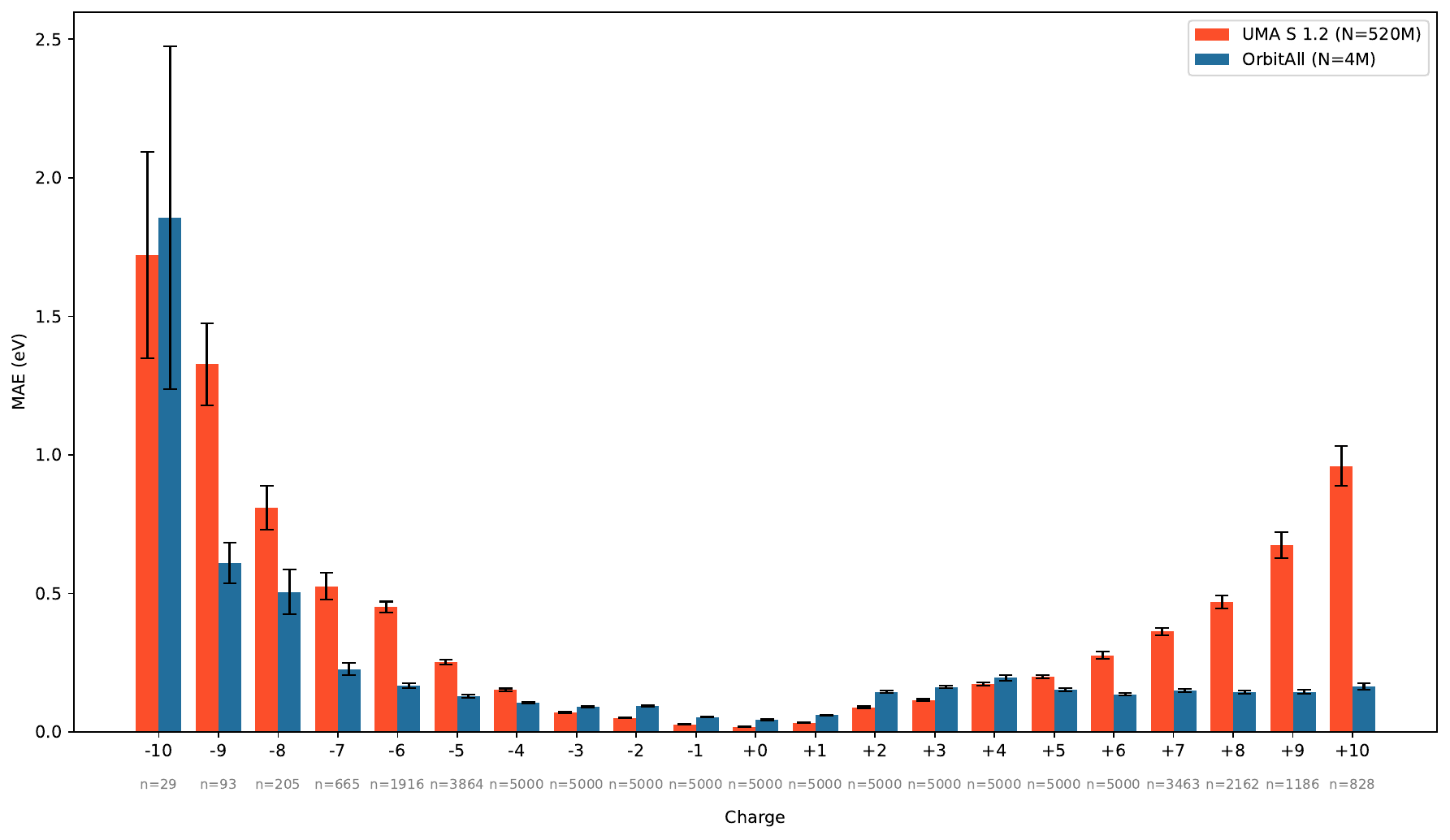}
    \caption{Validation-set MAE grouped by molecular charge, ranging from $-10$ to $+10$. The value below each charge label indicates the number of samples in that charge group, e.g., n=828 for charge $+10$.}
    \label{fig:app:omol_eval_by_charge}
\end{figure}

Figure \ref{fig:app:omol_eval_by_charge} shows the MAE across molecular charge states sampled from the validation set. Overall, \ourmodel{} achieves lower MAEs for highly charged species, particularly for cations with charges of +5 and above, and for anions with charges of -4 or below, except at charge -10. The improvement is especially clear for highly charged cations, where \ourmodel{} is more robust than UMA.

This trend may reflect the role of the g-xTB baseline and the resulting orbital features. For cations, g-xTB appears to provide a sufficiently accurate electronic-structure baseline, enabling \ourmodel{} to learn more accurate corrections. In contrast, anions are generally more challenging because their electron densities are more diffuse \cite{lynch2003anions_diffuse}, which can be difficult to represent with the minimal basis used in g-xTB. Nevertheless, the orbital features still appear to provide useful information for modeling highly charged species, contributing to the improved performance of \ourmodel{} across most extreme charge states.

Additionally, \ourmodel{} achieves MAEs of 34.03 meV for ``Biomolecules'', 69.50 meV for ``Electrolytes'', and 131.99 meV for ``Metal Complexes''. Although direct comparison is not straightforward because our training and evaluation sets differ from those used in prior OMol25-4M benchmarks (e.g., lanthanides are excluded in our setting), these errors are competitive with those reported for leading MLIPs trained on OMol25-4M. For example, GemNet-OC trained on OMol25-4M reports MAEs of 39.58 meV for ``Biomolecules'', 56.32 meV for ``Electrolytes'', and 148.49 meV for ``Metal Complexes''\cite{levine2025openmolecules2025omol25}. Notably, \ourmodel{} achieves lower errors for biomolecules and metal complexes, while remaining within a similar range for electrolytes. These results suggest that \ourmodel{} can model diverse and challenging chemical systems, including biomolecules, electrolytes, metal complexes, and non-equilibrium geometries, demonstrating its robustness and scalabil/ity.

\subsection{Solvation Reactions}

\begin{figure}[!ht]
    \centering
    \includegraphics[width=1\linewidth]{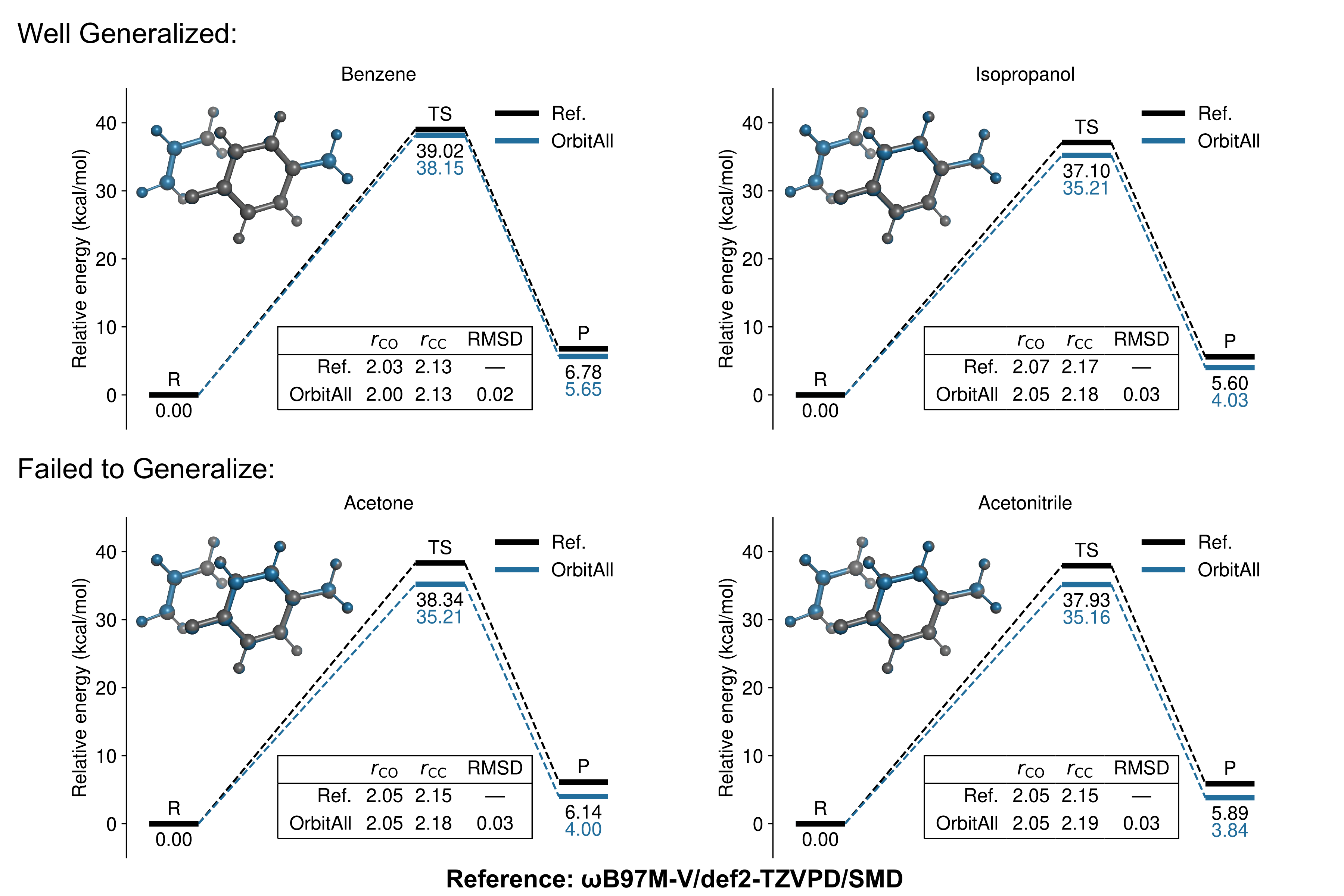}
    \caption{Evaluation of the generalization of \ourmodel{} to unseen solvent conditions. The top two panels, corresponding to benzene and isopropanol, show cases where \ourmodel{} generalizes well, whereas the bottom two panels, corresponding to acetone and acetonitrile, show cases where \ourmodel{} fails to generalize.}
    \label{fig:app:neb_other_solvs}
\end{figure}

Figure \ref{fig:app:neb_other_solvs} shows the zero-shot predictions of \ourmodel{} fine-tuned on the \tlxsolv{} dataset, which contains data in vacuum, water, methanol, and toluene. We observe two cases where the model generalizes successfully and two cases where it does not. The successful cases, benzene and isopropanol, are similar to solvents included in the training dataset. Benzene is similar to toluene, a nonpolar solvent without hydrogen-bonding capability, whereas isopropanol is similar to water and methanol, which are polar hydrogen-bonding solvents. In contrast, the two failure cases, acetone and acetonitrile, are both polar aprotic solvents that do not donate hydrogen bonds. SMD is a more sophisticated implicit solvation model than GBE, the base implicit solvation method used during the g-xTB data processing. In particular, SMD includes solvent-specific parameters that account for hydrogen-bonding effects, whereas GBE does not explicitly include such terms. Therefore, in future work, if the base method supports a more sophisticated implicit solvation model such as SMD, \ourmodel{} may generalize better to unseen solvents.

\begin{figure}[!ht]
    \centering
    \includegraphics[width=1\linewidth]{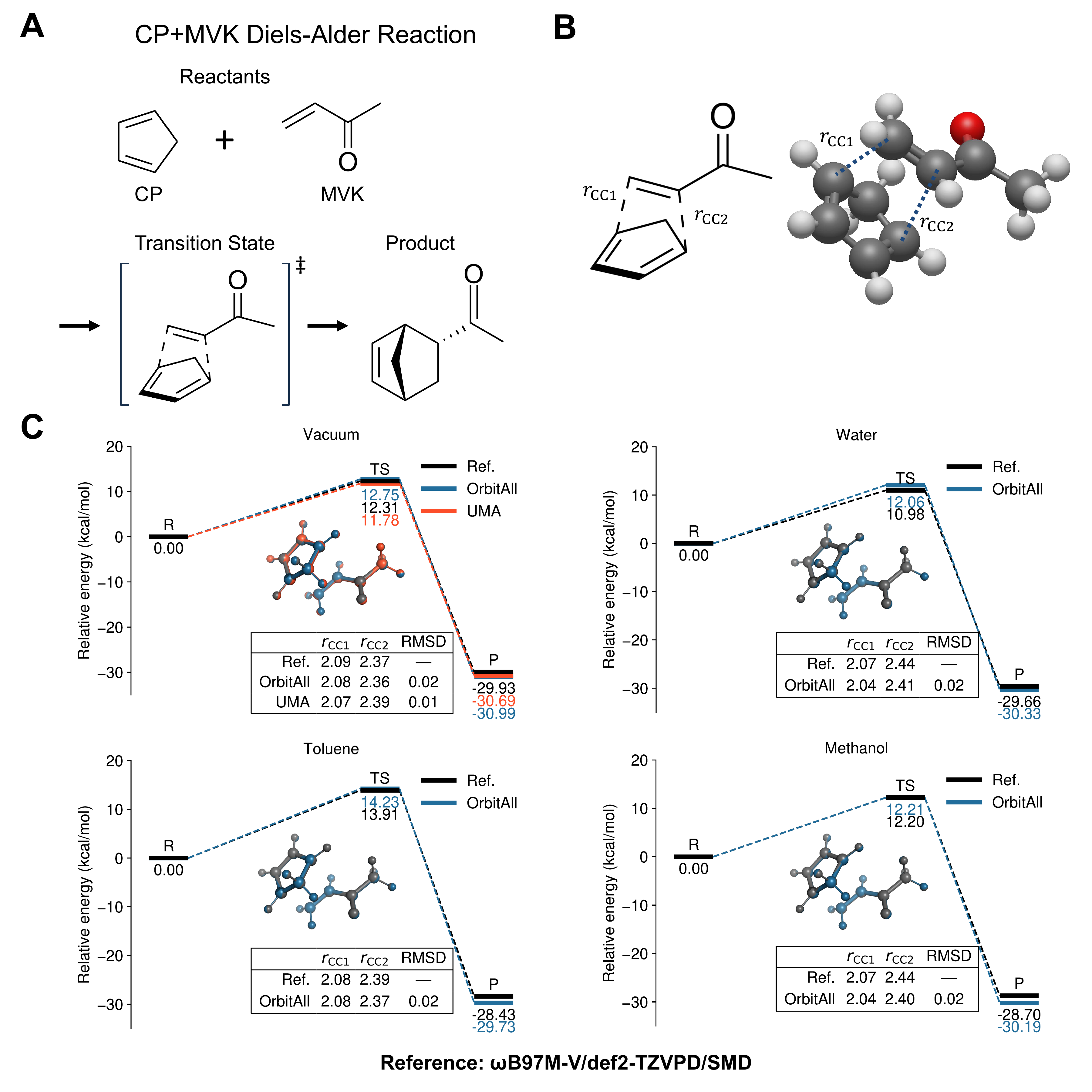}
    \caption{Transition-state (TS) search under implicit solvation for the cyclopentadiene (CP) + methyl vinyl ketone (MVK) Diels-Alder reaction. \textit{(A)}  Reaction schematic for the CP-MVK Diels-Alder reaction. \textit{(B)} Schematic and 3D example of the transition state, with the forming bond lengths ($r_\text{CC1}$ and $r_\text{CC2}$) labeled. \textit{(C)} NEB-based transition-state search for the CP-MVK Diels-Alder reaction for different solvents. ``Ref.'' denotes reference calculations at the $\omega$B97M-V/def2-TZVPD/SMD level of theory. For each solvent, the estimated reaction barriers (TS) and reaction energies (P) obtained from different methods are labeled below each state. Overlaid 3D transition-state structures from each method are shown, with Ref in gray, OrbitAll in blue, and UMA in orange. $r_\text{CC1}$, $r_\text{CC2}$, and the RMSD relative to the reference structure are reported in \AA, in the inset tables.}
    \label{fig:app:neb_da}
\end{figure}

We present additional results for transition state searches under implicit solvation for the Diels-Alder reaction between cyclopentadiene (CP) and methyl vinyl ketone (MVK) in Figure \ref{fig:app:neb_da}. Similar to the results shown in Figure \ref{fig:neb}, \ourmodel{} accurately predicts both the energetics and TS structures of the reaction under four conditions: vacuum, water, toluene, and methanol. All predicted energetics are near or within chemical accuracy (1 kcal/mol), and the TS structures are highly accurate, with RMSD values of only 0.02 \AA{} relative to the reference structures ($\omega$B97M-V/def2-TZVPD/SMD).

\ourmodel{} also accurately captures solvent effects. It correctly predicts that $r_\text{CC2}$ increases in water and methanol compared with vacuum and toluene. In terms of energetics, it also predicts that toluene increases the reaction barrier relative to vacuum, whereas water and methanol lower the barrier.

\section{Transition-state-informed Umbrella Sampling}

For umbrella sampling to be reliable, the sampled distributions of the collective variable from neighboring windows must exhibit sufficient overlap. We empirically find that a harmonic restraint force constant of $10~\mathrm{eV}/\text{\AA}^{2}$ provides good overlap between adjacent windows and enables stable sampling across the reaction coordinate. Representative window histograms for water are shown in Figure~\ref{fig:app:pmf_water_umbrella}, and the resulting potential-of-mean-force profiles along $\xi$ for different solvents are shown in Figure~\ref{fig:app:energetics_diff_solvents}.

\begin{figure}[ht]
\centering
\includegraphics[width=\linewidth]{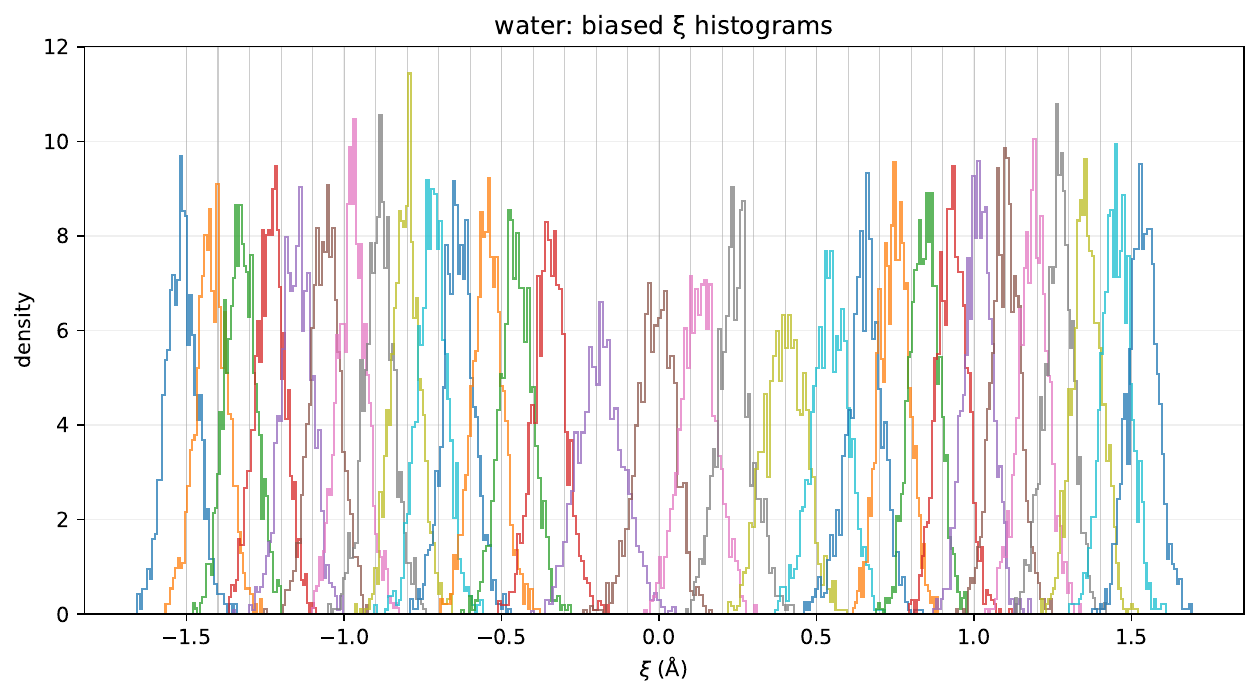}
\caption{Distribution histograms of the collective variable $\xi$ for the umbrella-sampling windows in water.}
\label{fig:app:pmf_water_umbrella}
\end{figure}

\begin{figure}[ht]
\centering
\includegraphics[width=\linewidth]{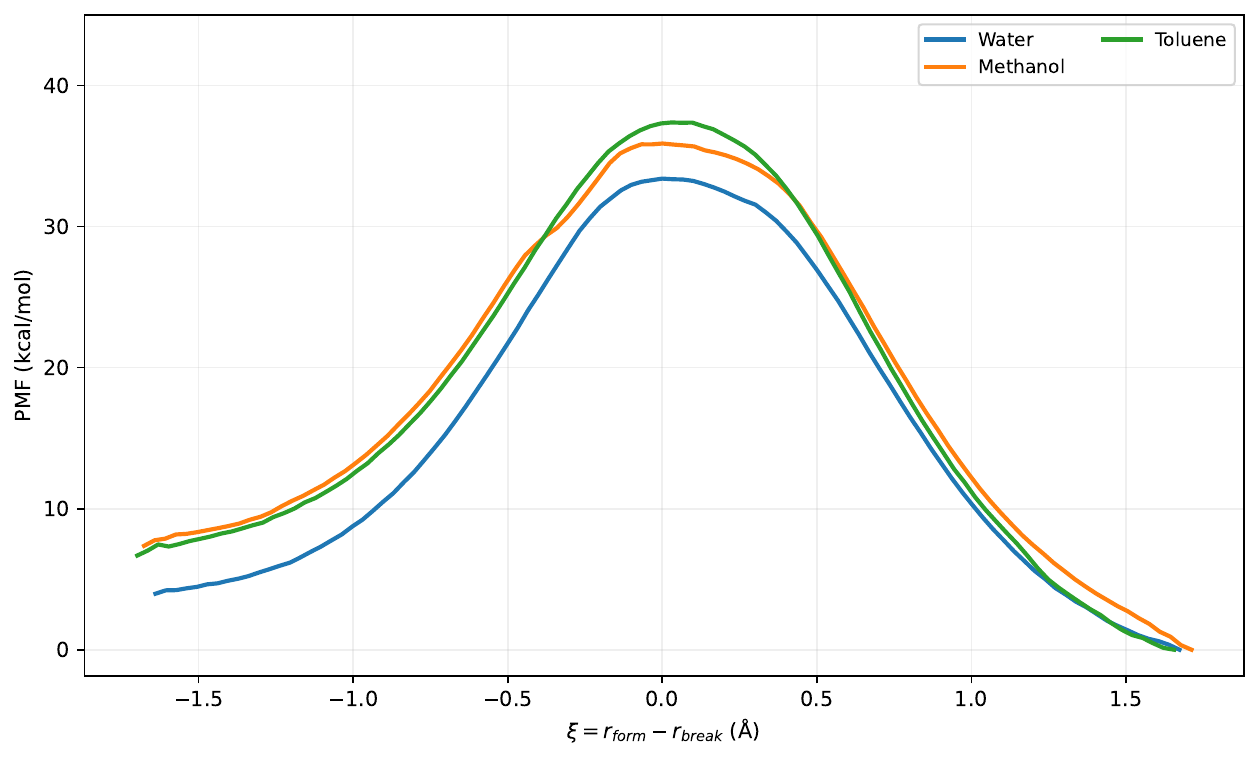}
\caption{Potential-of-mean-force profiles along the collective variable $\xi$ for different solvents.}
\label{fig:app:energetics_diff_solvents}
\end{figure}

\subsection{QM9star}

We present the numerical values of the experiments performed using the QM9star dataset in Table \ref{tab:app:qm9star_summary_table}.

\begin{table}[ht]
\setlength\tabcolsep{3pt}
\caption{Mean absolute error (MAE) in meV for the QM9star dataset. For the total energy $E_\text{tot}$ prediction task, the models were trained using the 400K training subset of the QM9star dataset, with 100K for each category of species. For HOMO and LUMO prediction tasks of radical species for $\alpha$ and $\beta$ spins, a 100K training set of radical species is used.}
\label{tab:app:qm9star_summary_table}
\begin{tabular}{Sc|ccccc|cccc|c}
& \multicolumn{5}{Sc|}{$E_\text{tot}$}                                                       & $E^\alpha_\text{HOMO}$ & $E^\alpha_\text{LUMO}$ & $E^\beta_\text{HOMO}$  & $E^\beta_\text{LUMO}$ & \multirow{2}{*}{Params} \\
& All           & Neutral       & Radical       & Cation         & Anion          & \multicolumn{4}{Sc|}{Radical}                                      &                             \\ \hline
\begin{tabular}[c]{@{}c@{}}SchNet\\ -SC(D)\end{tabular}     & 75.26     & 42.47        & 62.95       &  103.43        &  84.51        & -         & -       &-         & -        & 617K                        \\
\begin{tabular}[c]{@{}c@{}}PaiNN\\ -SC(D)\end{tabular}        &  41.63       &    23.40       &  34.67       & 56.66         &  47.51        & -       & -    & -        & -        & 723K                        \\
\begin{tabular}[c]{@{}c@{}}DimeNet++\\ -SC(D)\end{tabular}    &    19.41         &    7.46      &  14.42        &  30.34        &  22.59         & -      & -       & -         & -          & 1.4M                        \\
\begin{tabular}[c]{@{}c@{}}SphereNet\\ -SC(D)\end{tabular}    & 17.93         & 7.49          & 13.72         & 27.56          & 20.49        & -   & -         & -         & -        & 2.0M                        \\
\begin{tabular}[c]{@{}c@{}}EquiformerV2\\ -SC(D)\end{tabular} &  15.43        &   6.22 &  10.60        & 25.30        &  17.45      & -         & -          & -          & -         & 9.5M                        \\
\begin{tabular}[c]{@{}c@{}}NequIP\\ -SC(D)\end{tabular}       & 36.92     & 20.41         & 28.58         & 52.10         &  42.67       & -              & -              & -              & -              & 1.4M                        \\
OrbitAll(D)                                                   &  14.75 & 7.89        &  11.57 &   20.80 &    17.10   & - & - & - & - & 2.1M  \\ \hline
\begin{tabular}[c]{@{}c@{}}SchNet\\ -SC($\Delta$)\end{tabular}     & 33.31         & 18.57         & 29.15         & 42.68          & 39.39          & 151.16         & 129.43         & 79.19          & 57.36          & 617K                        \\
\begin{tabular}[c]{@{}c@{}}PaiNN\\ -SC($\Delta$)\end{tabular}        & 25.36         & 14.09         & 22.40         & 31.62          & 30.67          & 148.37         & 125.00         & 75.19          & 46.23          & 723K                        \\
\begin{tabular}[c]{@{}c@{}}DimeNet++\\ -SC($\Delta$)\end{tabular}    & 13.27         & 5.97          & 10.47         & 16.97          & 17.94          & 74.41          & 59.79          & 37.91          & 26.77          & 1.4M                        \\
\begin{tabular}[c]{@{}c@{}}SphereNet\\ -SC($\Delta$)\end{tabular}    & 12.94         & 5.65          & 10.22         & 16.69          & 17.49          & 90.15          & 80.18          & 49.22          & 30.93          & 2.0M                        \\
\begin{tabular}[c]{@{}c@{}}EquiformerV2\\ -SC($\Delta$)\end{tabular} & 9.86          & \textbf{3.44} & 7.13          & 13.66          & 13.70          & 65.96          & 49.38          & 38.66          & 23.59          & 9.5M                        \\
SpookyNet($\Delta$)                                                  & 19.27         & 10.76         & 15.43         & 23.97          & 24.93          & -              & -              & -              & -              & 3.7M                        \\
\begin{tabular}[c]{@{}c@{}}NequIP\\ -SC($\Delta$)\end{tabular}       & 23.06         & 12.78         & 19.73         & 29.22          & 28.09          & -              & -              & -              & -              & 1.4M                        \\
OrbitAll($\Delta$)                                                   & \textbf{8.50} & 4.86          & \textbf{6.89} & \textbf{10.60} & \textbf{10.78} & \textbf{55.40} & \textbf{43.31} & \textbf{32.27} & \textbf{14.73} & 2.1M                       
\end{tabular}
\end{table}

\subsection{QMSpin}
\label{app:qmspin_result}

\begin{table}[ht]
\small
\centering
\caption{Mean absolute error (MAE) in kcal/mol on total molecular energy when trained with 10,000 molecules in total on different training sets of the QMSpin dataset.}
\label{tab:app:singlets_vs_both}
\begin{tabular}{ccc}
\toprule
         & \ourmodel{} (Singlets Only) & \ourmodel{} (Singlets + Triplets) \\  \hline All      & 4.604                          & 0.786  \\
Singlets & 0.662                         & 0.819   \\
Triplets & 8.545                        & 0.753  \\ 
\hline
\end{tabular}
\end{table}

Table \ref{tab:app:singlets_vs_both} shows the results from training two \ourmodel{} models: (1) one using 10K singlets (Singlets-only) as a training set and 500 singlets as a validation set, and (2) one using 5K singlets and 5K triplets (Singlets + Triplets) as training set and 500 singlets and 500 triplets as a validation set, where all are selected randomly. We ensure geometries are not shared between training, validation, and test sets. Both models are tested against three sets: (1) a combined set of 2,316 remaining singlets and triplets (all), (2) the 2,316 singlets, and (3) the 2,316 triplets. We observe that \ourmodel{} can learn from the combined electronic structures of singlets and triplets. Specifically, the MAE is significantly improved when using both singlets and triplets as training data.

\begin{table}[ht]
\small
\centering

\caption{The mean absolute errors (MAEs) for the test set in meV on total molecular energy $E_\text{tot}$ when trained with 20K molecules for different models using the QMSpin dataset (MRCISD+Q-F12/cc-pVDZ-F12) \cite{schwilk2020qmspin}. The MAEs are taken from original works \cite{unke_spookynet_2021, simeon2025tensornet_spincharge, cheng_MOBML_2022}. OrbitAll and MOB-ML were trained on delta-labels \cite{cheng_MOBML_2022}. ``All'' is the combined test set of singlet ($S=0$) and triplet ($S=1$). SpookyNet \cite{unke_spookynet_2021} and TensorNet \cite{simeon2025tensornet_spincharge} reported ``all'' MAEs only. The best MAE for each category is highlighted with boldface.}
\label{tab:app:qmspin_comparison}
\renewcommand{\arraystretch}{1.25}

\begin{tabular}{Sccccc}
\hline
        & SpookyNet & TensorNet & MOB-ML & \textbf{OrbitAll (Ours)} \\ \hline
All     & 68.0                     & 43                          & 37.8                  & \textbf{28.5}     \\
Singlet ($E^{S=0}$) & -                          & -                                  & \textbf{27.0}          & 27.5          \\
Triplet ($E^{S=1}$) & -                          & -                                & 48.6                   & \textbf{29.5}     \\ \hline
$E^\text{gap}_\text{singlet}$ & - & - & 65.8 & \textbf{34.9} \\
$E^\text{gap}_\text{triplet}$ & - & - & 41.9 & \textbf{22.5} \\
$E^\text{gap}_\text{adiabatic}$ & - & - & 42.4 & \textbf{25.0} \\ \hline
\end{tabular}
\end{table}

We assess the performance of \ourmodel{} using the QMSpin dataset \cite{schwilk2020qmspin}, which contains open-shell species. The model predicts total energies of singlet ($E^{S=0}$) and triplet ($E^{S=1}$) carbenes at the MRCISD+Q-F12/cc-pVDZ-F12 level of theory. From these predictions, we compute vertical spin gaps at singlet-optimized ($E^\text{gap}_\text{singlet}$) and triplet-optimized ($E^\text{gap}_\text{triplet}$) geometries, as well as the adiabatic spin gaps ($E^\text{gap}_\text{adiabatic}$), as defined in \cite{cheng_MOBML_2022} as follows:

\begin{equation}
    \begin{split}
        &E^\text{gap}_\text{singlet}=E^{S=1}_\text{singlet}-E^{S=0}_\text{singlet},\\
    &E^\text{gap}_\text{triplet}=E^{S=1}_\text{triplet}-E^{S=0}_\text{triplet},\\
    &E^\text{gap}_\text{adiabatic}=E^{S=1}_\text{triplet}-E^{S=0}_\text{singlet},\\
    \end{split}
\end{equation}

\noindent
where the right-hand terms are total energies, with subscripts “singlet” and “triplet” indicating the multiplicity of the optimized geometry, and superscripts “$S=1$” and “$S=0$” indicating the spin state at which the energy is computed. The sampling strategy for the training, validation, and test sets is described in Section \ref{sec:datasets_and_trainings}.

As shown in Table \ref{tab:app:qmspin_comparison}, OrbitAll achieves the lowest MAEs across all categories except the singlet state, with both singlet and triplet errors smaller than chemical accuracy (1 kcal/mol $\approx$ 43.4 meV). This is significant given that the dataset was computed using the highly accurate but computationally demanding MRCISD+Q-F12/cc-pVDZ-F12 method. In contrast to MOB-ML, which trains separate models for singlet and triplet states using features from RHF and ROHF \cite{cheng_MOBML_2022}, OrbitAll employs a single model trained jointly on both, using the significantly cheaper spGFN1-xTB method. This unified training likely contributes to OrbitAll’s robust performance across spin states, whereas MOB-ML struggles with triplet predictions. Additionally, OrbitAll achieves the highest accuracy in predicting both vertical and adiabatic spin gaps, with all MAEs smaller than chemical accuracy. These results underscore the strength of \ourmodel{}’s unified representation in capturing and generalizing across different spin multiplicities.

\begin{figure}[!ht]
\begin{center}
\centerline{\includegraphics[width=1.0\columnwidth]{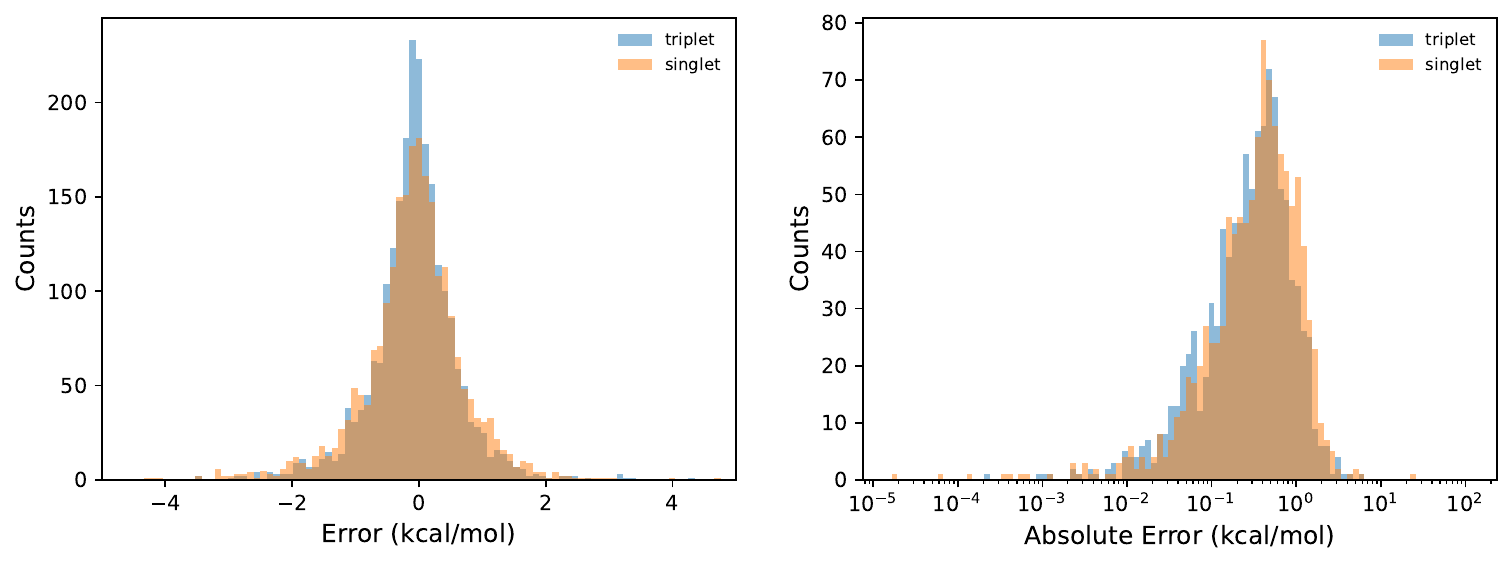}}

\caption{Error distribution histogram (left) and absolute error log distribution histogram (right) using \ourmodel{} trained with 10K singlets and 10K triplets, for predicting singlets and triplets in the test set of the QMSpin dataset.}

\label{fig:app:qmspin_error_histogram}
\end{center}
\end{figure}

Figure \ref{fig:app:qmspin_error_histogram} shows the error distribution histograms in terms of the errors and the absolute errors. The inference is done on the test set of the QMSpin dataset, with the \ourmodel{} model trained using 10K singlets and 10K triplets. We observe that \ourmodel{} is able to describe triplets and singlets almost equally well for the QMSpin dataset.

\subsection{Hessian QM9}

We additionally tested \ourmodel{} using different solvation methods in GFN1-xTB for generating orbital features. Specifically, we used the CPCM \cite{barone1998cpcm1, takano2005cpcm2}, ALPB \cite{sigalov2006alpb}, and GBSA \cite{qiu1997gbsa} implicit solvation methods implemented in the tblite package \cite{tblite}. For this test, we used a subset of the Hessian QM9 dataset split into 10K training, 1K validation, and 2K test molecules, with each molecule associated with four solvent data points: vacuum, toluene, THF, and water. For CPCM, we used the same dielectric constants as those used to generate the SMD-solvated data in Hessian QM9, whereas for ALPB and GBSA, we used the empirical solvent parameters implemented in tblite.

\begin{table}[ht]
\small
\centering
\caption{Mean absolute error (MAE) for the test set in kcal/mol for total molecular energy when trained on 10,000 molecules, corresponding to 40,000 data points across four solvents, from the Hessian QM9 dataset. The implicit solvation method used to generate the orbital features is indicated in parentheses.}
\label{tab:app:hessian_qm9_diff_solvents}
\begin{tabular}{cccc}
\toprule
Solvent & \ourmodel{} (CPCM) &  \ourmodel{} (GBSA) & \textbf{\ourmodel{} (ALPB)} \\
\hline
Vacuum & 0.367 &  0.350 & \textbf{0.328} \\
Toluene & 0.356  & 0.341  & \textbf{0.331} \\
THF & 0.361  & \textbf{0.330} & 0.335 \\
Water & 0.372  & \textbf{0.329} &  0.336 \\ \hline
All & 0.364 & 0.338 &  \textbf{0.332} \\
\hline
\end{tabular}
\end{table}

Interestingly, the results varied substantially depending on the implicit solvation method used to generate the orbital features. Based on this ablation study, we selected the ALPB method for evaluating \ourmodel{} on the Hessian QM9 dataset.

\subsection{Radical Species}

\begin{figure}[!ht]
    \centering
    \includegraphics[width=1\linewidth]{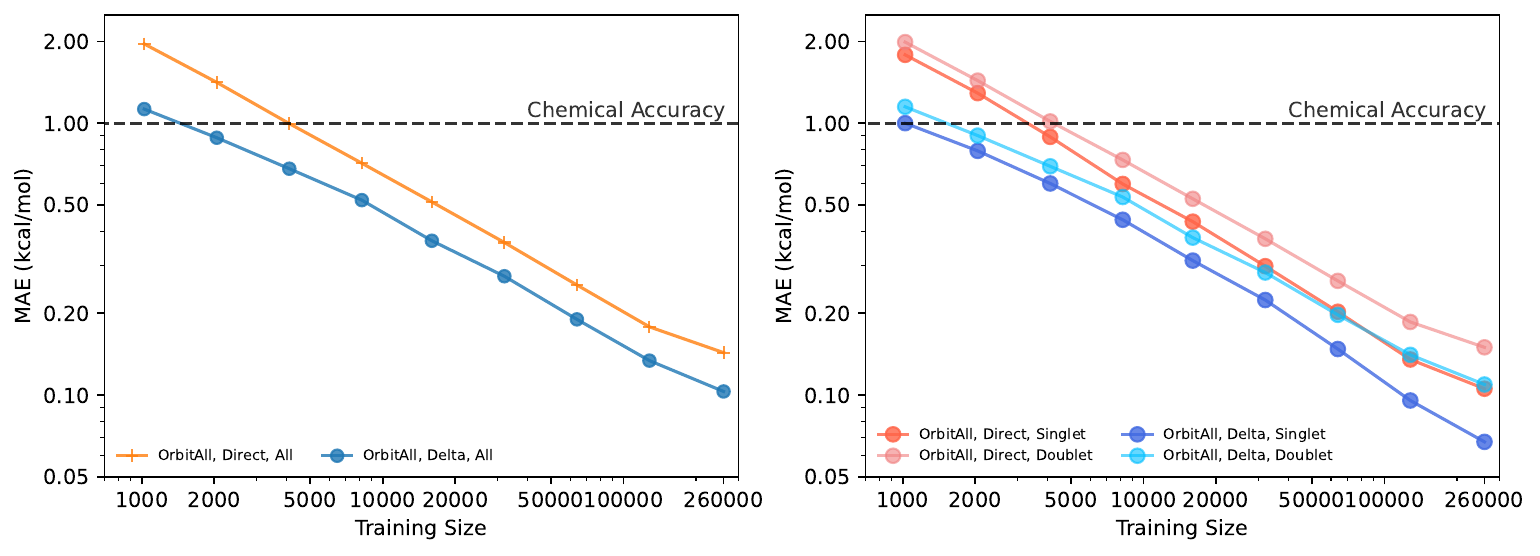}
    \caption{Learning curves of different models for the radicals dataset.}
    \label{fig:app:radicals_all_learning_curves}
\end{figure}

\begin{figure}[ht]
\begin{center}
\centerline{\includegraphics[width=1.0\columnwidth]{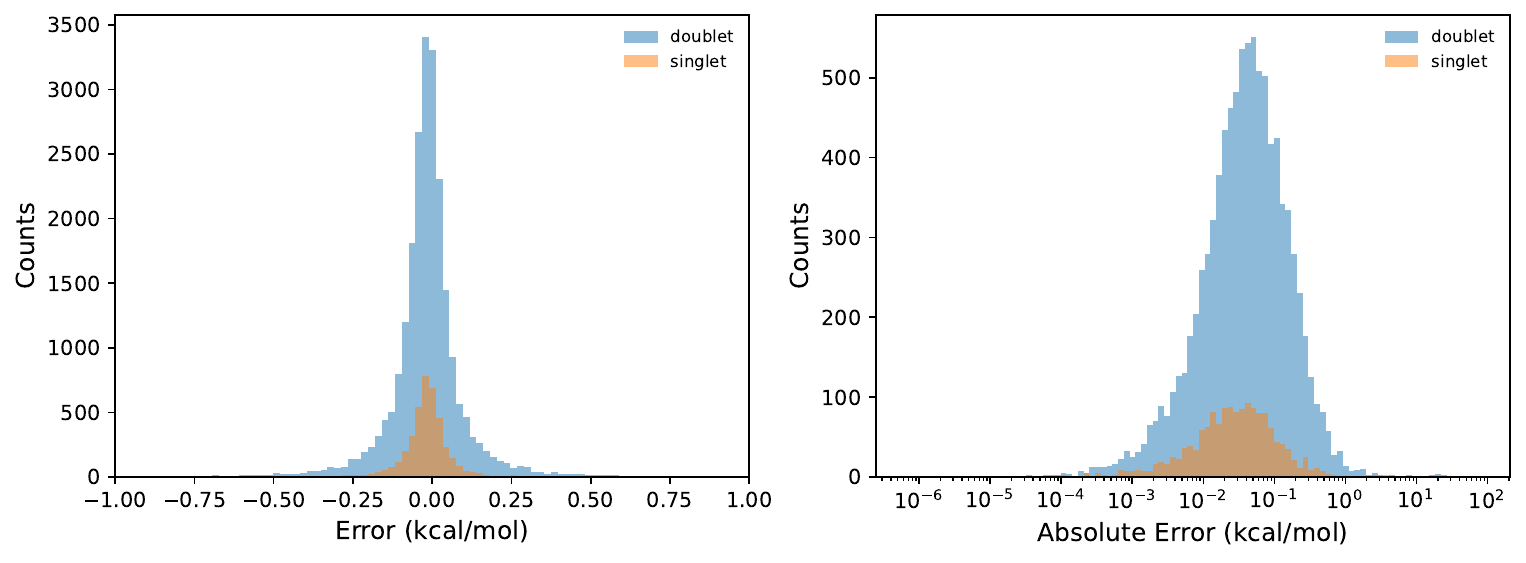}}

\caption{Error distribution histogram (left) and absolute error log distribution histogram (right) using \ourmodel{} trained with 260K molecules for predicting singlets and doublets in the test set of the radicals dataset from St. John et al. \cite{stjohn2020radicals_dataset}.}

\label{fig:app:radicals_error_histogram}
\end{center}
\end{figure}

In addition to the evaluations we presented in the main text, we also present the evaluation we performed on the dataset reported by St. John et al. \cite{stjohn2020radicals_dataset}. The dataset consists of $\sim$240K radicals and $\sim$40K closed-shell molecules. Specifically, we train on the SCF energies of each molecule for evaluation. Out of the entire dataset, we randomly sampled 260,000 molecules for training, 2,048 for validation, and the remaining 27,589 for testing. Then, we randomly sample a subset from the training set to create the learning curve.

The result implies that training for doublets is certainly more difficult than training for singlets. Although there were many more doublets in the dataset than singlets, we see that generally, singlets record lower MAEs at all training sizes.

\subsection{QM9star in Random Uniform External Electric Fields}
\label{app:qm9star_in_random_efield}

\begin{figure}[!ht]
    \centering
    \includegraphics[width=0.7\linewidth]{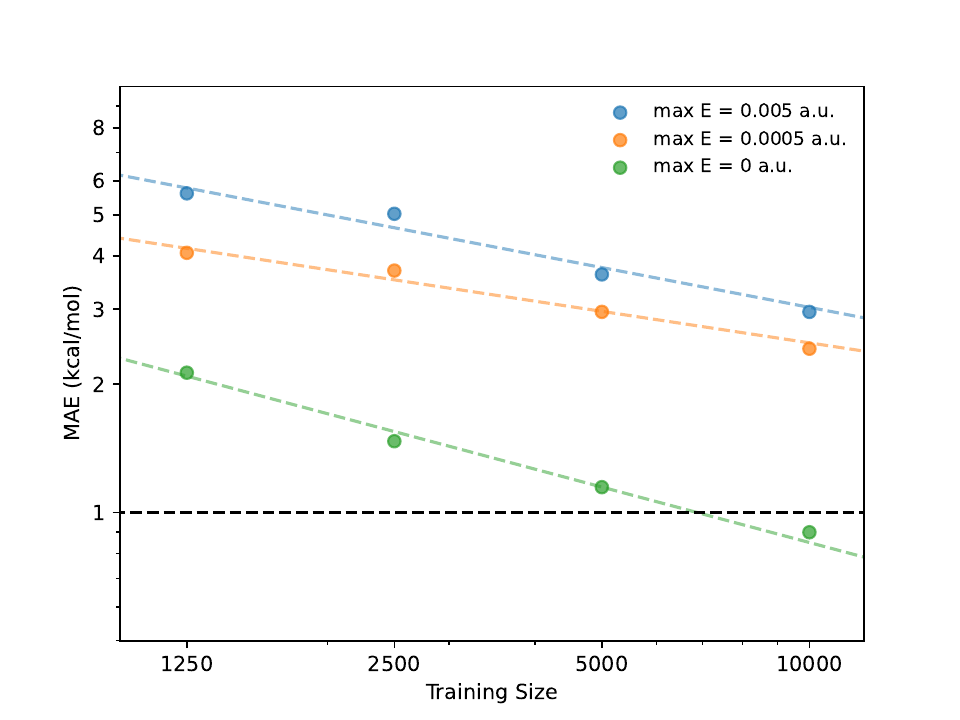}
    \caption{Learning curves of \ourmodel{} for different maximum magnitudes of uniform external electric fields.}
    \label{fig:app:qm9star_efield_learning_curves}
\end{figure}

Predicting the molecular energies perturbed by uniform external electric fields is observed to be significantly more challenging than those without external fields. In Figure \ref{fig:app:qm9star_efield_learning_curves}, we observe the higher offset and flatter slopes for both of the learning curves of different magnitudes of uniform external electric fields. The negative slopes for both cases of maximum magnitudes (0.005 au and 0.0005 au) imply that \ourmodel{} is capable of learning molecules in arbitrary uniform electric fields. However, the learning curves show that assuming the trend continues in a higher data regime, we require several hundred thousand data points to achieve chemical accuracy.

\subsection{Delta-learning}
\label{sec:delta_learning}

\begin{figure}[!ht]
    \centering
    \includegraphics[width=1\linewidth]{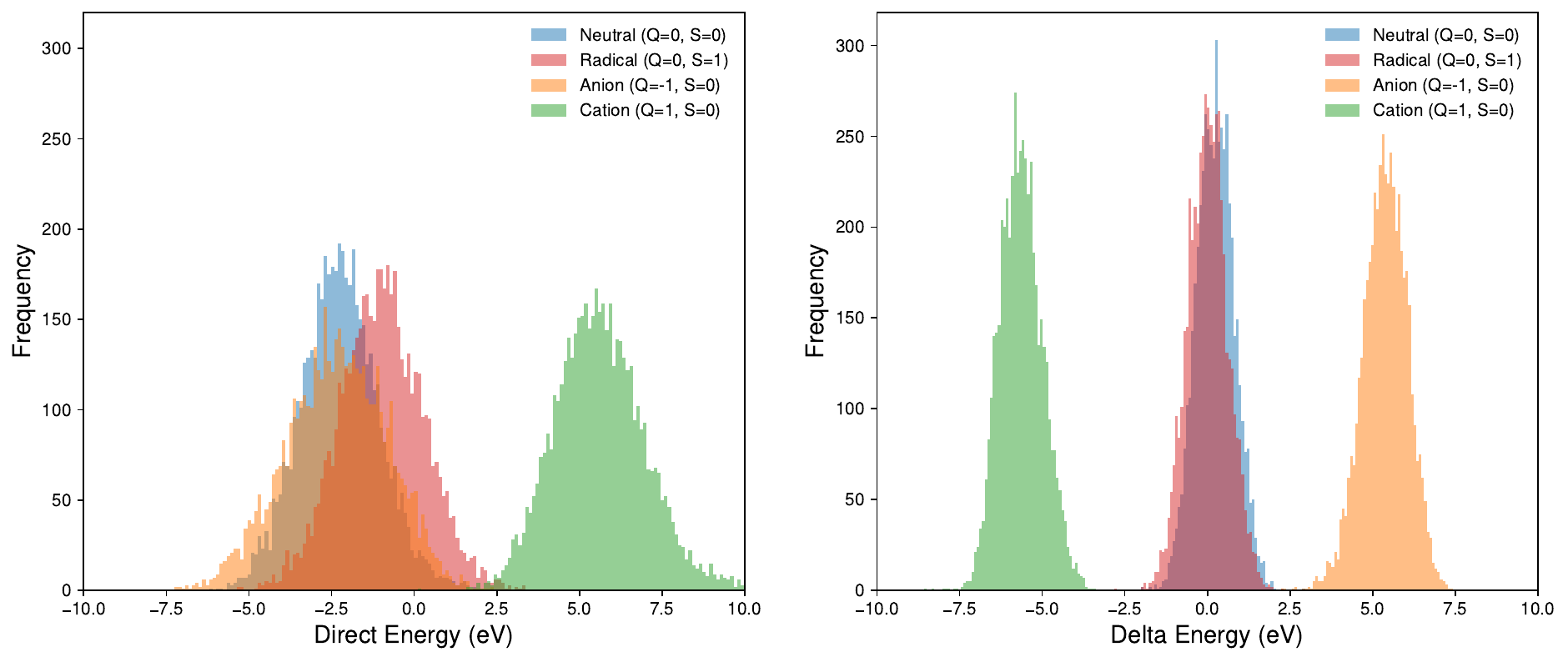}
    \caption{The label energy distribution histograms of different species for the QM9star dataset \cite{tang2024qm9star}, for direct-learning (left) and delta-learning (right). The delta-labels are obtained from subtractions by spGFN1-xTB energies \cite{neugebauer_spgfnxtb_2023}. The labels here are the original labels subtracted by the element-wise energy bias initializations.}
    \label{fig:app:qm9star_direct_delta_label_dist}
\end{figure}

Delta-learning is a widely used strategy in the area of chemical properties prediction tasks \cite{rama_delta_learning_2015}, which has been especially successful for energetic properties predictions \cite{ruth2022deltalearning1, chen2023deltalearning2, zhu2019deltalearning4}. Delta-learning is known empirically to reduce test error by a certain factor. Another study has shown that delta-learning can also account for long-range interactions \cite{böselt2021deltalearning3_lr}. Since accuracy also relates to data-efficient training, delta-learning can be particularly useful for highly expensive labels, such as energies at the CCSD(T) level of theory \cite{ruth2022deltalearning1}.

Figure \ref{fig:app:qm9star_direct_delta_label_dist} displays the data distribution pattern of different species in the QM9star dataset \cite{tang2024qm9star}. The distributions of direct-learning labels of different species are more dispersed. However, the distributions of the delta-learning labels appear to be more predictable and aligned, with sharper peaks around their corresponding means. Specifically, the anions and cations distributions become equidistant from the origin ($E=0$), where the center of distribution of the neutral species is located, with opposite signs. These changes in distributions facilitate generalizations, reducing systematic error patterns.

\section{Limitations and Future Works}

Our feature generation via the semi-empirical method in tblite does not support analytical gradients \cite{tblite}. Thus, \ourmodel{} cannot compute forces via backpropagation through atomic positions, unlike fully differentiable models. Obtaining forces from energy gradients ensures conservative force fields, which are necessary for stable geometry optimization and molecular dynamics \cite{bigi2025darkforcesassessingnonconservative}. Qiao et al. addressed this by computing numerical gradients of the intermediate features $\textbf{T}$ with respect to atomic positions, ${\partial\textbf{T}}/{\partial r_i}$ \cite{qiao_orbnet_equi_2022}.  However, this is computationally expensive and can lead to excessive memory usage if cached. Addressing this with analytical gradients \cite{qiao2020orbnetanalytical} or a differentiable semi-empirical backend \cite{friede_dxtb_2024} is possible in future work.

}
{}

\end{document}